%% file: main.tex
\documentclass[10pt,twocolumn,letterpaper]{article}

\usepackage[pagenumbers]{cvpr} 

\input{preamble}

\definecolor{cvprblue}{rgb}{0.21,0.49,0.74}
\definecolor{fbblue}{rgb}{0.25,0.34,0.57}
\definecolor{metablue}{rgb}{0,0.39,0.875}
\definecolor{umdred}{rgb}{0.887,0.097,0.199}
\usepackage[pagebackref,breaklinks,colorlinks,citecolor=cvprblue,linkcolor=umdred]{hyperref}
\usepackage{xcolor}
\usepackage{tabulary}
\usepackage{useful_math}
\usepackage{bbm}
\usepackage{colortbl}

\definecolor{myred}{RGB}{212,63,76}
\definecolor{mypurple}{RGB}{110,67,188}
\definecolor{mygray}{HTML}{868e96}

\newcommand{\ul}[1]{\underline{#1}}
\newcommand{\gray}[1]{\textcolor{mygray}{#1}}
\newcommand{\noindentnewline}{\\[4.5pt]\noindent}

\newcolumntype{x}[1]{>{\centering\arraybackslash}p{#1pt}}
\newcolumntype{y}[1]{>{\raggedright\arraybackslash}p{#1pt}}
\newcolumntype{z}[1]{>{\raggedleft\arraybackslash}p{#1pt}}
\newlength\savewidth\newcommand\shline{\noalign{\global\savewidth\arrayrulewidth
  \global\arrayrulewidth 1pt}\hline\noalign{\global\arrayrulewidth\savewidth}}
\newcommand{\tablestyle}[2]{\setlength{\tabcolsep}{#1}\renewcommand{\arraystretch}{#2}\centering\footnotesize}

\ExplSyntaxOn
\newcommand{\mytexttt}[1]{
  \small\texttt{
    \tl_set:Nn \l_tmpa_tl { #1 }
    \tl_replace_all:Nnn \l_tmpa_tl {~}{\hspace{0.3em}}
    \tl_use:N \l_tmpa_tl
  }
}
\ExplSyntaxOff

\def\paperID{1540}
\def\confName{CVPR}
\def\confYear{2024}

\title{
  Object Recognition as Next Token Prediction
  \vspace{-.3em}
}

\author{
    \hspace{-.82em}
    Kaiyu Yue$^{12}$\thanks{
      Work done during an internship at \href{https://ai.meta.com/}{\textcolor{black}{Meta AI}}.
      {\scriptsize\texttt{kaiyuyue@cs.umd.edu}}.
    }
    \hspace{.82em}  
    Bor-Chun Chen$^{1}$ \hspace{.82em} 
    Jonas Geiping$^{3}$ \hspace{.82em}  
    Hengduo Li$^{1}$ \hspace{.82em} 
    Tom Goldstein$^{2}$ \hspace{.82em} 
    Ser-Nam Lim$^{4}$
    \noindentnewline
    \hspace{-.9em}
    $^{1}$Meta
    \hspace{.3em}
    $^{2}$University of Maryland
    \hspace{.3em}
    $^{3}$ELLIS Institute \& MPI-IS T\"ubingen
    \hspace{.3em}
    $^{4}$University of Central Florida
}

\begin{document}
\maketitle
\input{sec/0_abstract}
\input{sec/1_intro}
\input{sec/4_related_works}
\input{sec/2_method}
\input{sec/3_exp}
\input{sec/X_suppl}
{
  \clearpage
  \small
  \bibliographystyle{ieeenat_fullname}
  \bibliography{main}
}
\end{document}

%% file: preamble.tex
\usepackage[dvipsnames]{xcolor}


%% file: sec/0_abstract.tex
\begin{abstract}
    \vspace{-.7em}
    \noindent
    We present an approach to pose object recognition as next token prediction.
    The idea is to apply a language decoder that auto-regressively predicts the text tokens from image embeddings to form labels.
    To ground this prediction process in auto-regression, we customize a non-causal attention mask for the decoder, incorporating two key features:
    modeling tokens from different labels to be independent, and treating image tokens as a prefix.
    This masking mechanism inspires an efficient method $-$ one-shot sampling $-$ to simultaneously sample tokens of multiple labels in parallel and rank generated labels by their probabilities during inference.
    To further enhance the efficiency, we propose a simple strategy to construct a compact decoder by simply discarding the intermediate blocks of a pretrained language model.
    This approach yields a decoder that matches the full model's performance while being notably more efficient.
    The code is available at \href{https://github.com/kaiyuyue/nxtp}{{\mytexttt{\textcolor{metablue}{github.com/kaiyuyue/nxtp}}}}.
\end{abstract}

%% file: sec/1_intro.tex
\vspace{-1.em}
\section{Introduction}
\label{sec:intro}
\looseness=-1
\noindent
This paper delves into a fundamental problem in computer vision $-$ object recognition $-$ translating an image into object labels.
Generally speaking, the recognition framework comprises an image encoder and a decoder.
The image encoder, either in the form of a convolutional neural network (CNN) \cite{liu2022convnet,he2016deep,krizhevsky2012imagenet,szegedy2015going,simonyan15vgg} or a vision transformer (ViT) \cite{vit,radford2021learning,xu2023metaclip}, produces image embeddings, while the decoder propagates them to predict object labels.
\noindentnewline
If the decoder is a linear classifier \cite{liu2022convnet,vit,he2016deep,krizhevsky2012imagenet,szegedy2015going,simonyan15vgg}, it needs to be initialized with fixed object concepts.
ResNet \cite{he2016deep}, for instance, initializes its final linear layer with 1K embeddings, a.k.a. weights, to represent 1K objects in ImageNet \cite{deng2009imagenet}.
Such static weights, however, limit the model's ability to recognize any object.
This limitation can be mitigated using a language model \cite{vaswani2017attention,devlin2018bert} as the decoder to generate a flexible set of object embeddings from input descriptions.
For example, CLIP \cite{radford2021learning} encodes the object descriptions into dynamic weights by prompting with ``a photo of a \{$\cL$\}'', where $\cL$ could be any object name,
and matches these weights with image embeddings to recognize objects.
\noindentnewline
Note that CLIP predefines the gallery with a fixed number of object descriptions prior to inference.
This requirement reveals that CLIP's object embeddings cover only a portion of the textual space in practical scenarios, rather than its entirety.
Additionally, enlarging the gallery has been shown to diminish its performance \cite{conti2023vocabulary}.
Given these observations, a question arises: Can we eliminate the predefined object labels or descriptions?
%
\begin{figure}[t]
  \centering
  \includegraphics[width=1\linewidth]{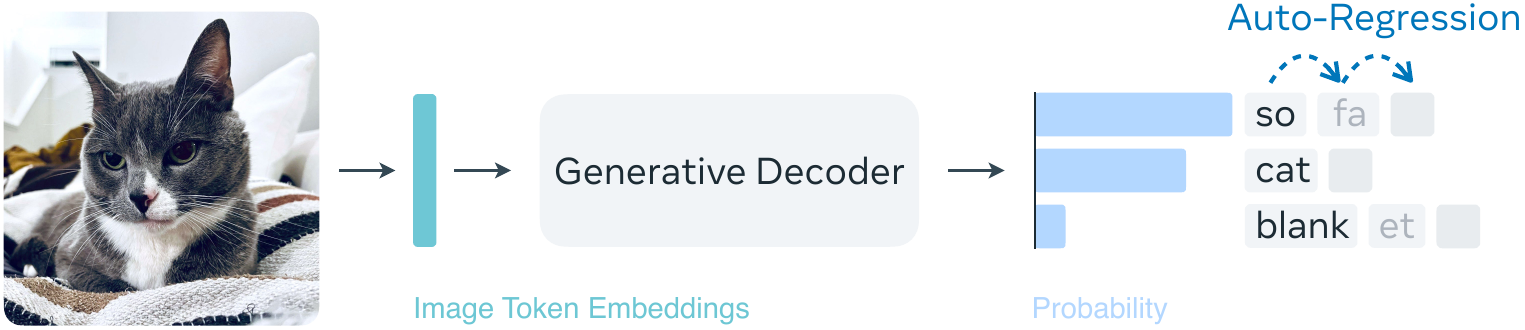}
  \vspace{-1.5em}
  \caption{
    \textbf{Object recognition as next token prediction} using a generative decoder such as a transformer-based language model to auto-regressively predict object labels.
      {\scriptsize{\textit{Photo authorized with \href{https://creativecommons.org/licenses/by/4.0/legalcode}{\textcolor{black}{CC BY 4.0}}.}}}
  }
  \label{fig:intro}
  \vspace{-1.5em}
\end{figure}
%
\noindentnewline
A direct strategy could use a generative model, particularly a large language model (LLM) \cite{vaswani2017attention,radford2018gpt1,radford2019gpt2,brown2020gpt3,gpt4,touvron2023llama,touvron2023llama2}, to decode labels from image embeddings.
For instance, Flamingo \cite{alayrac2022flamingo,awadalla2023openflamingo} employs a LLM to transform image embeddings into textual outputs for various vision tasks such as object recognition, image captioning, and visual question answering (VQA).
But producing the desired results for a specific task needs several reference samples as few-shot prompts for the model.
In other words, it requires predefined reference pivots to refine and align its predictions more precisely with the target task.
\noindentnewline
The most straightforward alternative is to skip any predefining procedure and align the LLM with the recognition task directly.
This approach hinges on the fact that a LLM's token embeddings represent the entire textual space, including all object labels.
This is as opposed to predefining subsets, i.e., query galleries or reference pivots, of this space that potentially constrains the model's capability.
\noindentnewline
\emph{Building on this concept, we propose a simple method that employs a language decoder to auto-regressively decode object labels token-by-token} from image embeddings, as depicted in Figure \ref{fig:intro}.
We operate a pretrained CLIP image encoder \cite{radford2021learning} to produce image embeddings, already aligned with text, and linearly transform them to match the language decoder's embedding dimension.
\noindentnewline
This auto-regressive framework, unlike the contrastive fra-mework exemplified by CLIP \cite{radford2021learning}, is trained to predict text embeddings from image embeddings, rather than aligning both.
While related in spirit to recent vision-language models such as LiMBeR \cite{merullo2022linearly}, LLaVA \cite{liu2023visual,liu2023improvedllava}, and BLIP-2 \cite{li2022blip,li2023blip2}, our method introduces differences and innovations:
\noindentnewline
First, our approach targets object recognition, as opposed to the chat-oriented VQA methods.
We train on image-caption pairs, easier to collect and annotate than image-question-answer triplets, and extract nouns from raw captions as reference labels to weakly supervise training.
For inference, we generate text fragments as labels rather than sentences.
In scenarios like recommendation systems \cite{resnick1997recommender} that require labels or tags, a simple label-only output is more concise than verbose sentences requiring further post-processing.
\noindentnewline
Second, our decoder has a different token modeling mechanism.
Instead of decoding all input and output tokens in a conditional sequence as in LLMs, we ensure tokens from different labels to be independent, while tokens from the same label remain conditional.
Naturally, all label tokens are conditional on image embeddings.
This decoupling is based on the understanding that different labels in the same image are independent but their coexistence is determined by the underlying visual context.
To this end, we customize a non-causal attention mask for our language decoder.
\noindentnewline
Further, the non-causal masking mechanism inspires a new sampling method, called \emph{one-shot sampling}, to generate text tokens for labels.
Instead of sampling tokens in sequence as in greedy search, beam search, and nucleus sampling \cite{holtzman2020curious}, one-shot sampling simultaneously samples tokens of multiple labels in parallel and ranks them by their probabilities.
This makes use of the strong parallelization capabilities of a transformer, leading to object recognition that is much more efficient than the aforementioned methods and does not suffer from repetition issues \cite{fu2021theoretical,xu2022learning}.
\noindentnewline
Lastly, we put forth a straightforward strategy to enhance model efficiency of our recognition model.
We hypothesize that only partial knowledge in LLMs is vital for recognition and focus on maximizing efficiency by not engaging the entire language model.
To construct the decoder, we start with a pretrained LLM, e.g., LLaMA \cite{touvron2023llama,touvron2023llama2}, retain first six transformer blocks along with the final output layer, and drop the intervening blocks.
This compact decoder matches the full model's performance but is substantially more efficient, i.e., 4.5$\times$ faster in inference.

%% file: sec/4_related_works.tex
\section{Related Work}
\label{sec:related works}
\looseness=-1
\noindent
\textbf{Aligning Images and Text}, including sentences, phrases, or words, in a shared space has been prevalent for image-text matching \cite{hofmann2001plsa,blei2003lda,de2006generating,weston2011wsabie,frome2013devise,socher2014grounded,kiros2014multimodal,karpathy2014deep,mao2014explain}, and foundational in contrastive frameworks \cite{gupta2020contrastive,radford2021learning,ma2023world}, while others are geared towards generating text descriptions from images \cite{mao2014explain,socher2014grounded,kiros2014multimodal,karpathy2015deep,showandtell,johnson2016densecap}.
Then, integrating visual perception with LLMs \cite{vaswani2017attention} like GPT \cite{radford2018gpt1,radford2019gpt2,brown2020gpt3,gpt4} and LLaMA \cite{touvron2023llama,touvron2023llama2} is gaining traction by treating image embeddings as language token embeddings, seamlessly fusing visual and textual information within the model \cite{lstm,sherstinsky2020fundamentals}.
Such methods are being applied to tasks such as detection \cite{chen2021pix2seq}, few-shot recognition \cite{radford2021learning,alayrac2022flamingo}, textual explainations \cite{borth2013large}, classification justification \cite{hendricks2016generating}, bottleneck models \cite{yang2023bottle,saifullah2023seeing}, reasoning \cite{andreas2016reasoning,hendricks2018grounding,menon2022visual,han2022zero,mao2023doubly,schwettmann2023multimodal}, and chat-based models \cite{merullo2022linearly,liu2023visual,liu2023improvedllava,li2022blip,li2023blip2,dai2023instructblip} for captioning and VQA.
\noindentnewline
\textbf{Tackling Open-Vocabulary Tasks} for recognition \cite{radford2021learning}, detection \cite{zareian2021open,gu2021open,minderer2022simple,du2022learning,kuo2022f,minderer2023scaling} and segmentation \cite{du2022learning,ghiasi2022scaling} typically involves training on a set of base labels and then recognizing rare unseen labels.
The cornerstone of open-vocab approaches is the contrastive learning \cite{hadsell2006dimensionality,sohn2016improved} like CLIP \cite{radford2021learning}, which employs a language model to encode labels to contrast with images.
Therefore, open-vocab methods potentially inherit CLIP's limitations discussed in Section \ref{sec:intro} due to the predefined base and rare labels.
CaSED \cite{conti2023vocabulary} utilizes raw captions to form a vocabulary-free gallery, diverging from the gallery of predefined label vocabularies.
However, its performance is heavily dependent on gallery selection, as demonstrated in Table 10 of \cite{conti2023vocabulary}, highlighting its limitations as a retrieval-based method.
\noindentnewline
We argue that by dramatically increasing the training data to cover a wide array of objects, the reliance on recognizing rare data and concepts can be heavily reduced.
Our method aligns more with the open-world paradigm \cite{bendale2015towards} that incrementally learns new labels over time, mirroring the way of data collection in the real world.
In the application, given just an image, our model predicts labels with ranking probabilities, without relying on any predefined set of concepts.

%% file: sec/2_method.tex
\section{Method}
\label{sec:method}
\subsection{Revisiting Object Recognition}
\noindent
We begin by briefly reviewing object recognition in its general formulation.
Suppose that 2D images are fed into a backbone, e.g. ViT \cite{vit} in CLIP \cite{radford2021learning}, which produces image embeddings\footnote{
  Bold capital letters denote a matrix $\X$, and bold lower-case letters a column vector $\x$.
  $\x_i$ and $\x^j$ represents the $i^{\text{th}}$ row and $j^{\text{th}}$ column of the matrix $\X$ respectively.
  $\X_{ij}$ denotes the scalar in the $i^{\text{th}}$ row and $j^{\text{th}}$ column of the matrix $\X$.
  All non-bold letters represent scalars.
} $\X_\text{v} \in \real^{M \times D}$, where $M$ is the spatial size and $D$ is the embedding dimension.
In a nutshell, the problem of recognition aims to decode object labels solely from $\X_\text{v}$, translating image embeddings into the textual space.
\noindentnewline
In the past years, the core design of this translation employs a set of textual embeddings $\W \in \real^{N \times D}$ to seek the optimal alignment with $\X_\text{v}$:
\begin{aligns}
  \label{eq:recognition}
  \argmax \ \sigma(\W f(\X_{\text{v}})^\top),
\end{aligns}
where $\sigma$ is the softmax function and $f$ is to transform $\X_\text{v}$ for aligning with $\W$.
For instance, linear classifiers such as ResNet \cite{he2016deep} employ the average pooling as $f$ to transform $\X_\text{v}$ to a single vector representation, and initiate $\W$ using a set of predefined concepts corresponding to object labels, e.g., $N=1000$ for ImageNet \cite{deng2009imagenet}.
The contrastive frameworks such as CLIP \cite{radford2021learning} embed a collection of predefined object descriptions into $\W$, and apply an aggregation (like [CLS] embedding \cite{vit}) and linear projection as $f$ on $\X_\text{v}$.
\noindentnewline
Eq. \ref{eq:recognition} aims to maximize the alignment between $f(\X_\text{v})$ and $\W$.
The space of $\W$ plays a critical role in this alignment as the diversity and richness of the embeddings in $\W$ directly affect the model's ability to differentiate objects.
The linear classifiers and contrastive frameworks, however, limit $\W$ to a predefined subset that potentially constrains the model's capability to recognize any object.
Our goal is to eliminate this limitation and extend $\W$ to the entire textual space.
\subsection{Auto-Regression for Recognition}
\begin{figure}[t]
  \centering
  \includegraphics[width=1.\linewidth]{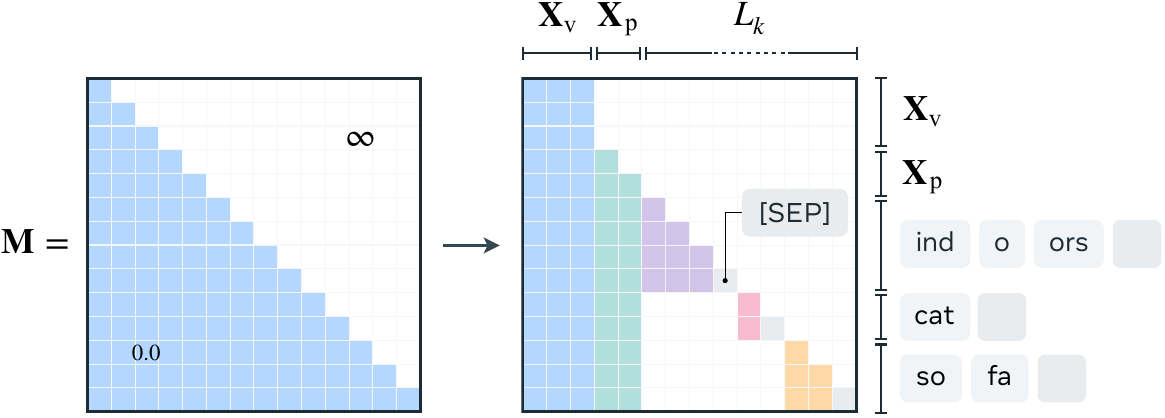}
  \vspace{-1.5em}
  \caption{
    \textbf{Non-causal attention mask} for prefixing image tokens $\X_{\text{v}}$ and decoupling tokens from different labels $L_k$ to be independent at the [SEP] token.
  }
  \label{fig:non-causal-attention-mask}
  \vspace{-1.5em}
\end{figure}

\noindent
Recently, LLMs have significantly advanced in understanding and generating text \cite{vaswani2017attention,radford2018gpt1,radford2019gpt2,brown2020gpt3,gpt4,touvron2023llama,touvron2023llama2}.
Considering that their token embeddings are trained to represent the entire textual space, we define $\W$ with the token embeddings\footnote{
  In general, LLMs have two sets of token embeddings, one for encoding input tokens and the other for predicting output tokens.
  Some LLMs like GPT-2 \cite{radford2019gpt2} share the same embeddings for both input and output tokens \cite{press2016using}, while others like LLaMA \cite{touvron2023llama2} employ different embeddings.
  Our method defines $\W$ with the embeddings designated for output tokens.
} from a pretrained LLM, e.g., LLaMA \cite{touvron2023llama,touvron2023llama2}, featuring $N=32$K textual tokens.
Then Eq. \ref{eq:recognition} changes to predicting the token:
\begin{aligns}
  \label{eq:recognition-token-prediction}
  P(\w | \X_\text{v}) = \argmax \ \sigma(\W f(\X_{\text{v}})^\top),
\end{aligns}
where $\w$ represents the most probable single token for $\X_\text{v}$.
In our method, $f$ is a combination of linear projection and the pretrained LLM to project $\X_\text{v}$ in the textual space of $\W$.
That is, $f$ is our language decoder.
\noindentnewline
To guide the language decoder in the recognition task, we prompt it with a short instruction $-$ ``the objects in the image are'' $-$ tokenized as $\X_\text{p} \in \real^{P \times D}$.
Then we concatenate $\X_{\text{v}}$ and $\X_{\text{p}}$ to form our input token embeddings:
\begin{aligns}
  \X = \X_\text{v} \oplus \text{[IMG]} \oplus \X_\text{p},
\end{aligns}
where $\oplus$ is the concatenation operation and [IMG] is a special token to indicate the boundary.
\noindentnewline
Typically, a label consists of multiple tokens, e.g., ``sofa'' has two tokens [so] and [fa].
Without loss of generality, we assume a label $L$ has $T$ tokens.
Now predicting $L$ is equivalent to auto-regressively predicting its tokens:
\begin{aligns}
  \label{eq:recognition-token-prediction-auto-regressive}
  P(L) = P(\w_1, \dots, \w_T | \X_\text{v}, \X_\text{p}) = \prod_{t=1}^{T} P(\w_t | \w_{<t}, \X),
\end{aligns}
where $\w_t$ is the $t$-th token of $L$, and $\w_{<t}$ is the sequence of tokens before the $t$-th token.
To compute the conditional probability in Eq. \ref{eq:recognition-token-prediction-auto-regressive}, the transformer-based LLM in $f$ employs a causal mask $\M$ \cite{vaswani2017attention} on the pairwise attention $\A$ to model the interdependence between tokens:
\begin{aligns}
  \A \leftarrow \A + \M, \quad \M = \text{tril}(\infty),
\end{aligns}
where $\text{tril}(\infty)$ is with zeros in the lower triangle and infinity values in the upper triangle.
This enforces the token $\w_t$ to attend only to the preceding tokens $\w_{<t}$, i.e., making $\w_t$ conditional on $\w_{<t}$, as shown in the left of Figure \ref{fig:non-causal-attention-mask}.
%
%
\subsection{Non-causal Masking}
\label{sec:non-causal-masking}
\noindent
In general, an image contains multiple objects, and our goal is to predict them all.
Suppose there are $K$ objects, and we denote the output set of labels for the image as $\mathcal{L} = \{L_1, ..., L_K\}$, where $k$-th label has $T_k + 1$ tokens, including the special token [SEP] for the delimiter.
Then the likelihood of this set of labels appearing in the image is the product of their probabilities:
\begin{aligns}
  \label{eq:multi-label}
  P(\mathcal{L}) = \prod_{k=1}^{K} P(L_k) = \prod_{k=1}^{K} \prod_{t=1}^{T_k + 1} P(\w^k_t | \w^k_{<t}, \X).
\end{aligns}
Now Eq. \ref{eq:multi-label} is not a standard auto-regression practiced in LLMs because $\w^k_t$ only needs to attend to the input tokens $\X$ and the preceding tokens $\w^k_{<t}$ from the same label $L_k$.
This is supported by the understanding that the labels coexist in the same image due to the underlying visual context, but are independent of each other.
Additionally, the image tokens $\X_\text{v}$ exhibit inherently spatial correlation, in contrast to the temporal correlation of natural language tokens.
Therefore, we customize a non-causal attention mask $\M$ with two designs, illustrated in the right of Figure \ref{fig:non-causal-attention-mask}:
a) We decouple the correlation between tokens from different labels at the [SEP] token to prevent these tokens from being attended to each other;
b) We treat image tokens $\X_\text{v}$ as a prefix \cite{liu2018generating,raffel2020exploring,wang2022language,wang2021simvlm,diao2022write,wang2022git}, enabling the image tokens to \emph{see} each other.
Interestingly, our non-causal attention mask shares a similar design as the column mask in \cite{ramesh2021zero} but is developed from a different perspective, where the column mask is specifically for image-to-image attention.
\noindentnewline
In the end, Eq. \ref{eq:multi-label} is our final training objective.
We use the cross-entropy loss for optimization, with weakly-supervised labels\footnote{
  Our learning approach is considered weakly-supervised as the labels are incomplete and imperfect derived from raw captions.
} $\cL$ extracted from the corresponding image captions.
\subsection{One-Shot Sampling}
\label{sec:one-shot-sampling}
\begin{figure}[h]
  \vspace{-.5em}
  \centering
  \includegraphics[width=1.\linewidth]{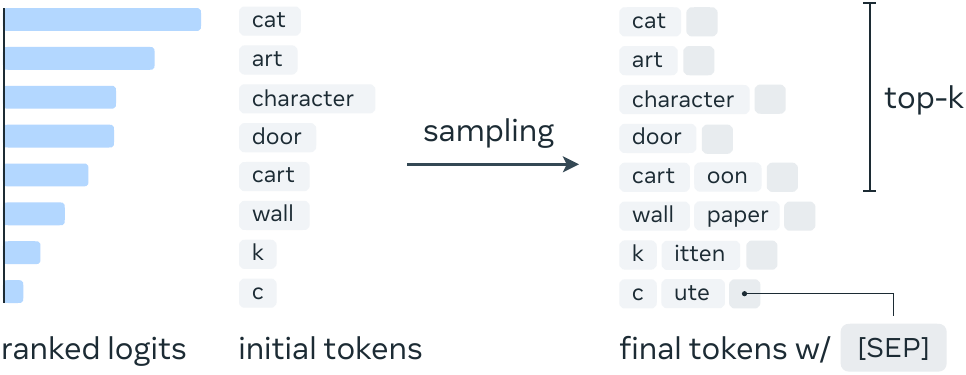}
  \vspace{-1.5em}
  \caption{
    \textbf{One-shot sampling} for generating tokens of top-$k$ labels in parallel.
    Once the model samples the [SEP] token, the label is completed.
    Otherwise, the model continues for unfinished labels.
  }
  \label{fig:one_shot_sampling}
  \vspace{-1.em}
\end{figure}

\noindent
The non-causal masking decouples the tokens from distinct labels, indicating that the first token of any label could be the next after $\X$ in the first sampling round.
In other words, a higher probability for the first token, being sampled after input $\X$, would result in a higher relevance of the label to the image.
This inspires us to sample tokens of multiple labels in parallel, as shown in Figure \ref{fig:one_shot_sampling}.
%
%
\noindentnewline
Given input tokens $\X$, we propagate them into the decoder and rank the output logits by their softmax probabilities.
The top-$k$ tokens, called initial tokens, decide the top-$k$ labels to be generated.
The efficacy of linking initial tokens to final labels is explored in Table \ref{tab:results_ablation_ranking_one_shot}, highlighting the promise of this straightforward approach.
Then we sample the next token for the top-$k$ initial tokens in parallel, using top-$1$ sampling, to generate $k$ labels.
If the sampled token is [SEP], the label is completed.
Otherwise, the model continues to sample the next token for the unfinished labels.
Finally, we report the probability of each label as the product of its token probabilities.
We refer to this approach as \emph{one-shot sampling,} which enables parallel sampling of multiple labels in one shot.
The key to its parallelism lies in the non-causal masking mechanism, which also avoids the repetition issue \cite{fu2021theoretical,xu2022learning} typically faced in greedy and beam search, as it causes the model to focus uniformly on the same input tokens $\X$ across various labels.
\noindentnewline
To sum up, the one-shot sampling differs from other sampling methods in two essential aspects:
a) It operates in parallel across multiple object labels, with each parallel branch processing a small number of tokens (roughly less than ten tokens), in contrast to the sequential sampling of other methods;
b) It naturally aligns with the vision recognition task by representing the image as a spatially correlated entity, while other sampling methods depict the image as a sequence of tokens.
\subsection{Truncating the Decoder}
\label{sec:truncating_decoder}
\begin{figure}[h]
  \vspace{-.5em}
  \centering
  \includegraphics[width=1.\linewidth]{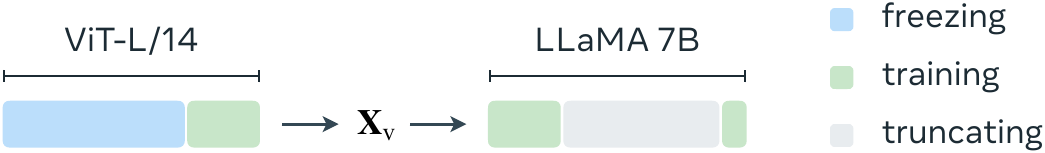}
  \vspace{-1.5em}
  \caption{
    \textbf{Encoder and truncated decoder}.
    We retain the first 6 transformer blocks along with the final output layer of the LLaMA 7B as our truncated decoder, and train with partial encoder blocks.
  }
  \label{fig:implementation}
  \vspace{-1.em}
\end{figure}

\noindent
Now, considering the language model LLaMA in our decoder $f$, we posit that a specific subset of language understanding in its numerous parameters is vital for recognition.
This realization prompts us to focus on maximizing efficiency by not engaging the entire model.
We construct our language decoder, initially based on the LLaMA 7B (version 1 or 2), by truncating it to the first 6 transformer blocks along with the final output layer, as depicted in Figure \ref{fig:implementation}, while preserving its tokenizer and the pretrained 32K token embeddings for encoding the input.
We designate this modified version as the truncated language decoder, denoted as Lang$_{\text{truncated}}$ in our experiments.
%
%

%% file: sec/3_exp.tex
\section{Experiments}
\label{sec:exp}
\noindent
\textbf{Data}.
We construct training datasets at two different scales for experiments.
\textbf{G3M}: a training group of 3M(illion) pairs combines CC3M \cite{sharma2018conceptual}, COCO Captions \cite{lin2014microsoft,chen2015microsoft}, SBU \cite{ordonez2011im2text}, which is mainly used for ablation studies.
\textbf{G70M}: We gather 67M pairs from LAION-Synthetic-115M (slightly fewer than previous work due to missing URLs) \cite{li2022blip,schuhmann2021laion}.
Combining it with G3M, we form a 70M-pair training group for scaling-up training.
For evaluation, we use the validation split of CC3M, COCO Captions, and OpenImages V7 \cite{benenson2022colouring}.
We parse the \textit{raw captions} to obtain meaningful nouns as reference labels in both training and evaluation.
The processing details are described in Section \ref{sec:x_data_preprocessing}.
\noindentnewline
\textbf{Implementation}.
The inference augmentation for input images in CLIP \cite{radford2021learning} is applied in both training and evaluation.
The input size is $224^2$.
The image encoder is ViT-L/14 \cite{vit} pretrained from CLIP \cite{radford2021learning}, producing 256 token embeddings with the dimension of 1024, as $\X_\text{v}$.
Note that we drop its [CLS] token.
The special token embedding of [IMG] is learned during training.
The special token [SEP] is the comma (,), and 32K token embeddings for the input are fixed.
The max number of input tokens is 512.
No [EOS] token, i.e., the end of the sentence, is used in the input.
We shuffle labels for each image in training.
\noindentnewline
\textbf{Training}.
AdamW \cite{loshchilov2017decoupled} with the cosine annealing learning rate (LR) schedule \cite{loshchilov2016sgdr} is applied in single-stage training.
The multi-dimensional parameters apply a weight decay of $10^{-1}$.
The global batch size is 512 with 32 NVIDIA A100-SXM4-80GB GPUs.
The warm-up has 2K iterations.
We jointly train four parts:
the last 6 blocks of the image encoder ViT-L/14,
the projection layer for $\X_\text{v}$,
the special [IMG] token embedding,
and the whole truncated language decoder, using a LR of $10^{-5}$ for 3 epochs, as shown in Figure \ref{fig:implementation}, taking {\small{$\sim$}}5 hours on G3M and {\small{$\sim$}}5 days on G70M.
\noindentnewline
\textbf{Evaluation}.
The $n$-gram overlap metrics, including BLEU \cite{papineni2002bleu} and ROUGE \cite{lin2004rouge}, are widely used to evaluate the quality of sentences generated by language models.
However, these metrics are not suitable for evaluating the quality of results in recognition tasks.
For example, ``car'' and ``automobile'' have the low $n$-gram similarity but are semantically alike.
To quantify the semantic similarity between the generated labels and the reference labels, we adopt the concept from BERTScore \cite{zhang2019bertscore} to formulate our evaluation metric\footnote{
    The metric essentially measures the model's accuracy,
    as explained in Section \ref{sec:x_evaluation_metric}.
}.
\noindentnewline
Formally, given a set of reference labels $\cR$ with size $M$ and a set of generated labels $\cG$ with size $N$,
we use the sentence-BERT \cite{reimers2019sentence} to encode $\cR$ to a set of semantic embeddings $\R \in \real^{M \times D}$ and $\cG$ to $\G \in \real^{N \times D}$, where $D$ is the embedding dimension.
Then we compute the cosine similarity matrix $\bS \in \real^{M \times N}$ between $\R$ and $\G$:
\begin{aligns}
    \label{eq:cosine_similarity}
    \bS_{ij} = \frac{\br_{i} \ \g_{j}^\top}{\|\br_i\| \|\g_{j}\|} \in \real^{[-1, 1]}.
\end{aligns}
We compute the recall for the reference set $\R$ and the precision for the generated set $\G$:
\begin{aligns}
    \label{eq:pr}
    R = \frac{1}{M} \sum_{i=1}^M \max_{j} \bS_{ij},
    \quad
    P = \frac{1}{N} \sum_{j=1}^N \max_{i} \bS_{ij},
\end{aligns}
where $\max$ indicates the greedy matching strategy following \cite{zhang2019bertscore}.
Finally, we compute the $F_1$ score as the harmonic mean of $R$ and $P$:
\begin{aligns}
    F_1 = \frac{2RP}{R+P}.
\end{aligns}
For each sample, we evaluate the top-$k$ generated labels out of $N$ and report the average $R$, $P$, and $F_1$ over all samples.
\noindentnewline
Note that, different models may have different numbers of generated labels $N$ for each image.
Especially, when $N < k$, we do not pad the matrix $\bS$ with zeros to make $N = k$ and penalize the model.
Thus, the model with $N < k$ will have a higher $P$ compared to the model with $N = k$.
\subsection{Main Results}
\label{sec:main_results}
%
\begin{figure}[t]
    \vspace{.1em}
    \centering
    \begin{minipage}{0.5\linewidth}
        \includegraphics[width=\linewidth]{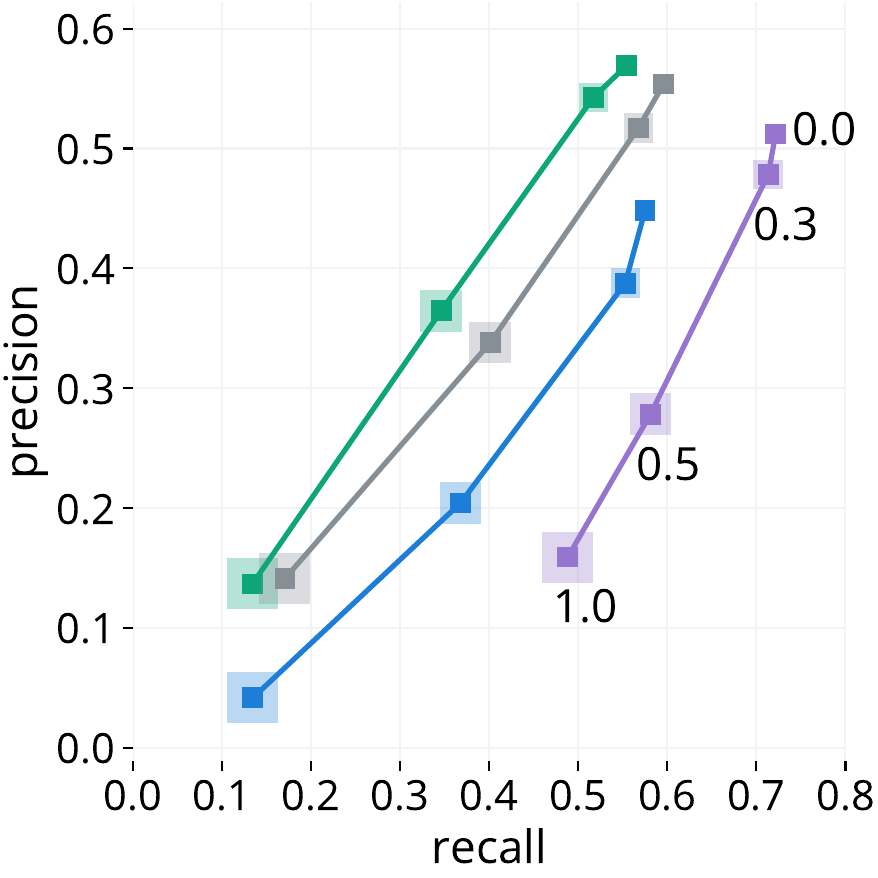}
    \end{minipage}%
    \begin{minipage}{0.5\linewidth}
        \includegraphics[width=\linewidth]{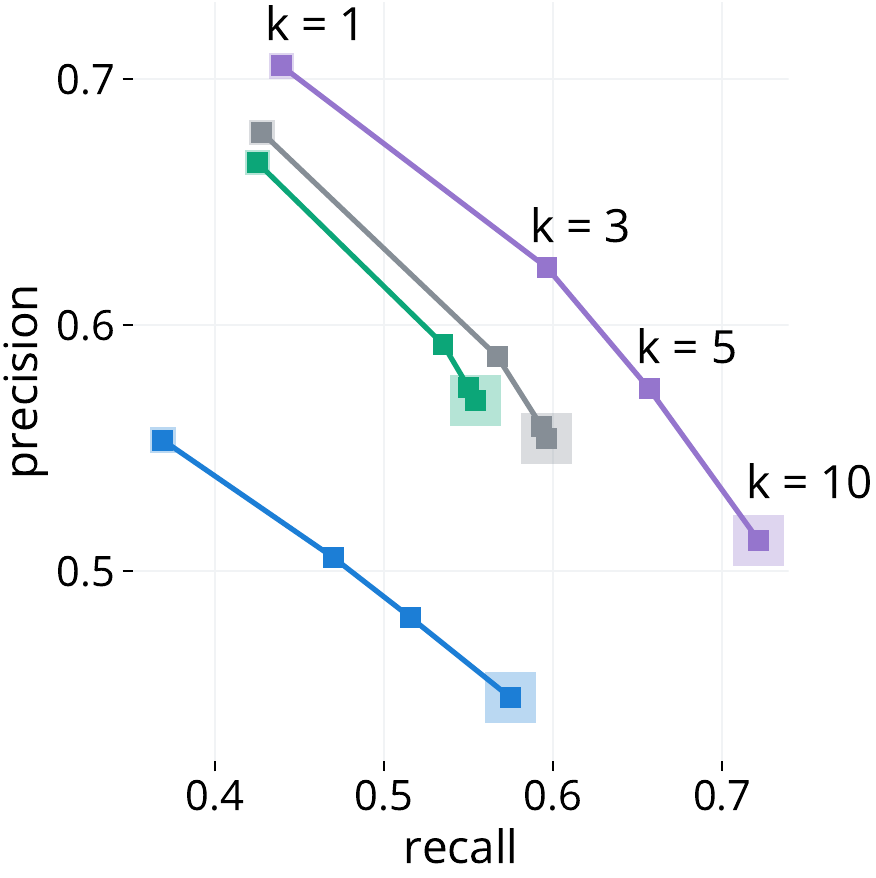}
    \end{minipage}%
    \\[3.pt]
    \begin{minipage}{0.5\linewidth}
        \includegraphics[width=\linewidth]{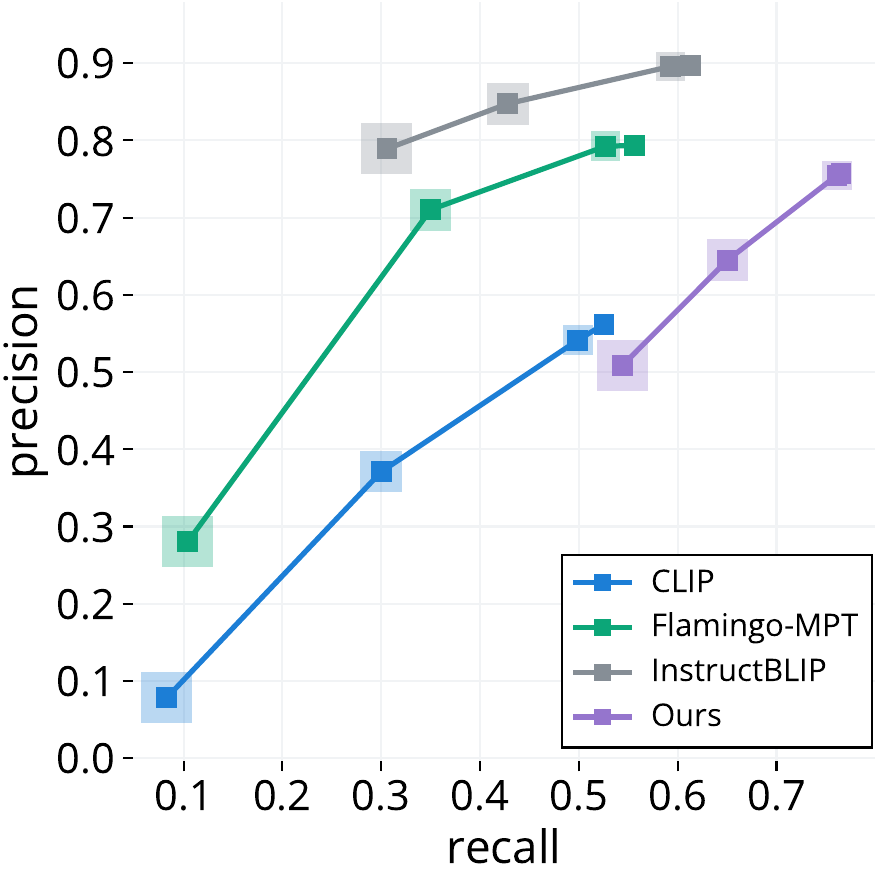}
    \end{minipage}%
    \begin{minipage}{0.5\linewidth}
        \includegraphics[width=\linewidth]{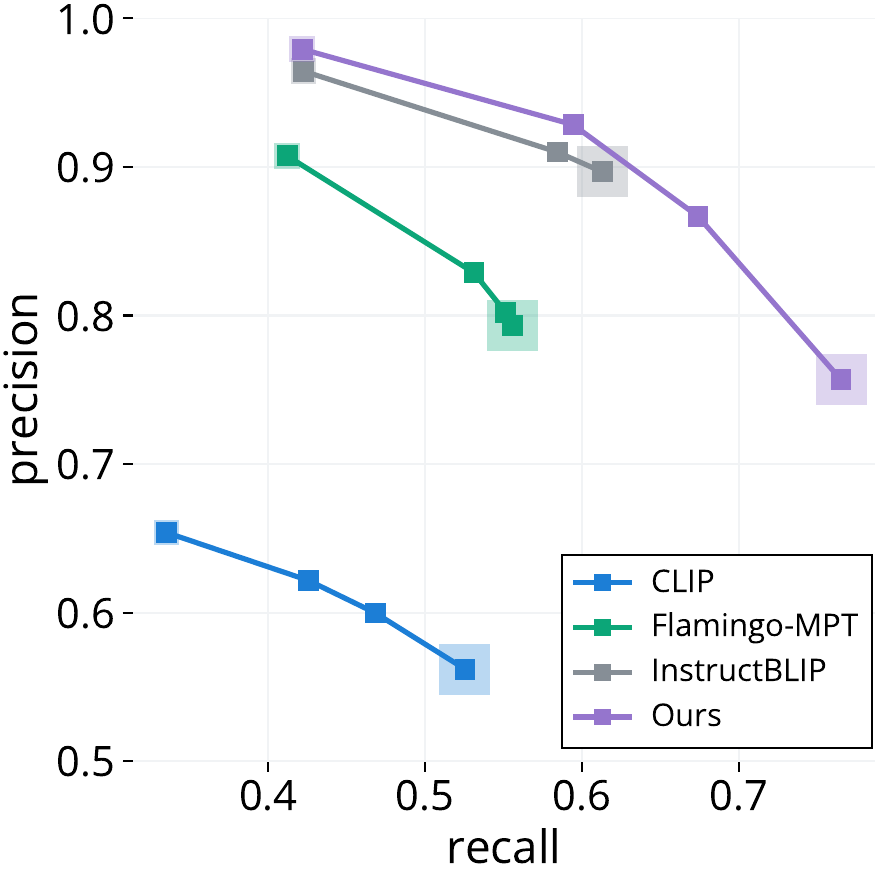}
    \end{minipage}%
    \\[3.pt]
    \begin{minipage}{0.5\linewidth}
        \includegraphics[width=\linewidth]{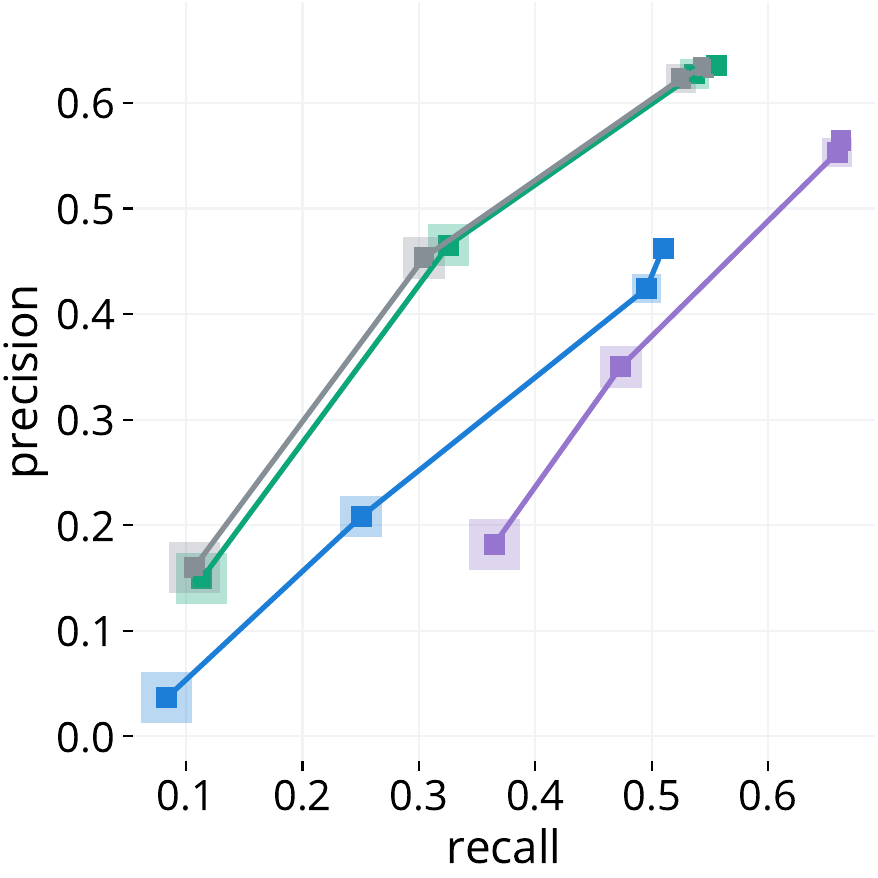}
    \end{minipage}%
    \begin{minipage}{0.5\linewidth}
        \includegraphics[width=\linewidth]{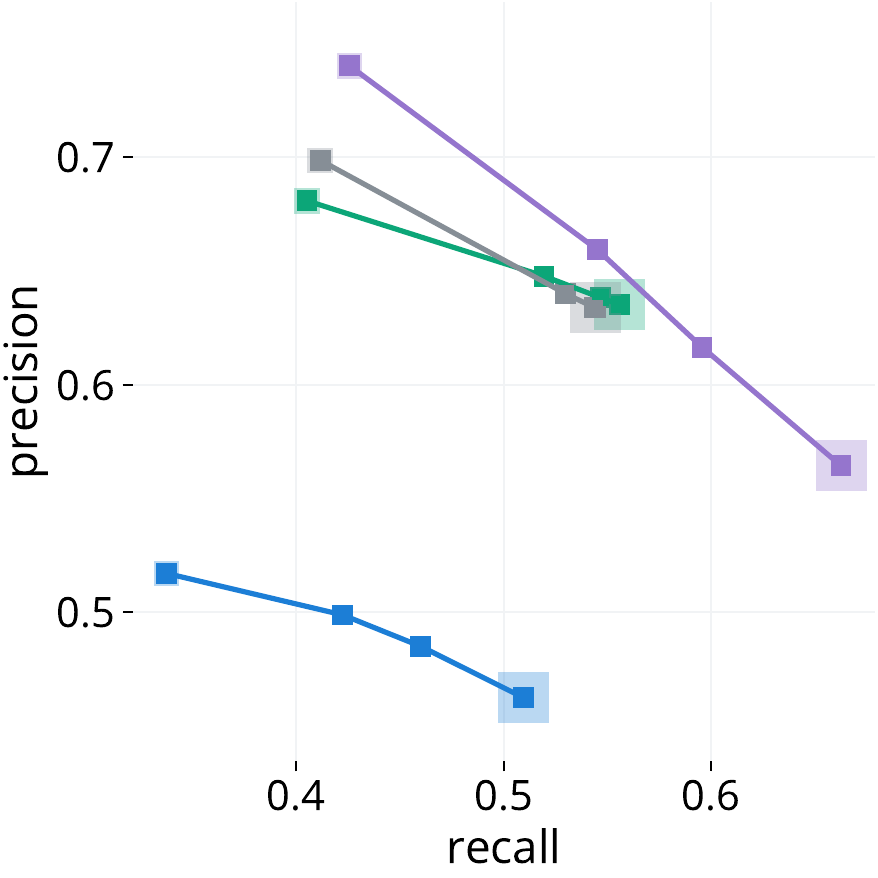}
    \end{minipage}%
    \vspace{-.5em}
    \caption{
        \textbf{Precision-recall (PR) curves} on CC3M, COCO, and OpenImages validation splits within 3 rows from top to bottom.
        The left column is the PR curves with different thresholds, i.e., $[0.0, 0.3, 0.5, 1.0]$, applying on the similarity matrix $\bS$ in Eq. \ref{eq:cosine_similarity}.
        The right column is the PR curves with different top-$k$ predictions, where $k$ is $[1, 3, 5, 10]$.
        All figures share the same legend.
    }
    \label{fig:pr_curves}
    \vspace{-1.em}
\end{figure}
%
%
\begin{table*}[t]
    \vspace{-2.em}
    \centering
    \begin{minipage}{1.\linewidth}
        \tablestyle{2.4pt}{1.15}
        \begin{tabular}{y{57}|y{94}|x{28}|z{36}|x{46}|x{18.5}x{18.5}x{18.5}|x{18.5}x{18.5}x{18.5}|x{18.5}x{18.5}x{17.5}}
                                      &                                                 &                &                        &               &
            \multicolumn{3}{c|}{CC3M} &
            \multicolumn{3}{c|}{COCO} &
            \multicolumn{3}{c}{OpenImages}                                                                                                                                \\
            method
                                      & \ models (vision + lang)                        & prompt         & data scale  \ \        & \# params (B)
                                      & R                                               & P              & F$_1$
                                      & R                                               & P              & F$_1$
                                      & R                                               & P              & F$_1$                                                          \\
            \shline
            \gray{CLIP} \cite{radford2021learning}
                                      & \ \gray{ViT L-14 + CLIP$_{\text{lang}}$}        & \gray{-}       & \gray{400M} \ \ \ \ \  & \gray{0.43}
                                      & \gray{0.575}                                    & \gray{0.448}   & \gray{0.499}
                                      & \gray{0.525}                                    & \gray{0.562}   & \gray{0.540}
                                      & \gray{0.510}                                    & \gray{0.462}   & \gray{0.480}                                                   \\
            \gray{CaSED} \cite{conti2023vocabulary}
                                      & \ \gray{ViT L-14 + Retrieval}                   & \gray{-}       & \gray{12M} \ \ \ \ \   & \gray{0.43}
                                      & \gray{0.648}                                    & \gray{0.471}   & \gray{0.540}
                                      & \gray{0.582}                                    & \gray{0.592}   & \gray{0.584}
                                      & \gray{0.534}                                    & \gray{0.470}   & \gray{0.494}                                                   \\
            \hline
            CLIP \cite{radford2021learning}
                                      & \ ViT L-14 + CLIP$_{\text{lang}}$               & -              & 400M \ \ \ \ \         & 0.43
                                      & 0.451                                           & 0.383          & 0.409
                                      & 0.429                                           & 0.483          & 0.450
                                      & 0.386                                           & 0.363          & 0.371                                                          \\
            CaSED \cite{conti2023vocabulary}
                                      & \ ViT L-14 + Retrieval                          & -              & 403M \ \ \ \ \         & 0.43          & 0.653 & 0.481 & 0.548
                                      & 0.616                                           & 0.629          & 0.620
                                      & 0.560                                           & 0.494          & 0.519                                                          \\
            Flamingo$_{\text{open}}$ \cite{awadalla2023openflamingo}
                                      & \ ViT L-14 + LLaMA 1 \cite{touvron2023llama}    & list           & 2.1B \ \ \ \ \         & 8.34
                                      & 0.547                                           & 0.540          & 0.536
                                      & 0.549                                           & 0.721          & 0.618
                                      & 0.526                                           & 0.621          & 0.562                                                          \\
            Flamingo$_{\text{open}}$
                                      & \ ViT L-14 + LLaMA 1                            & caption        & 2.1B \ \ \ \ \         & 8.34
                                      & 0.548                                           & 0.521          & 0.527
                                      & 0.553                                           & 0.697          & 0.611
                                      & 0.538                                           & 0.607          & 0.563                                                          \\
            Flamingo$_{\text{open}}$
                                      & \ ViT L-14 + MPT \cite{MosaicML2023Introducing} & list           & 2.1B \ \ \ \ \         & 8.13
                                      & 0.554                                           & \ul{0.569}     & 0.553
                                      & 0.556                                           & 0.793          & 0.646
                                      & 0.555                                           & 0.635          & 0.584                                                          \\
            Flamingo$_{\text{open}}$
                                      & \ ViT L-14 + MPT                                & caption        & 2.1B \ \ \ \ \         & 8.13
                                      & 0.534                                           & 0.533          & 0.527
                                      & 0.554                                           & 0.754          & 0.633
                                      & 0.551                                           & 0.613          & 0.574                                                          \\
            LLaVA$_\text{1.0}$ \cite{liu2023visual}
                                      & \ ViT L-14 + LLaMA 2 \cite{touvron2023llama2}   & list           & 753K \ \ \ \ \         & 13.3
                                      & 0.540                                           & 0.528          & 0.526
                                      & 0.580                                           & 0.803          & 0.666
                                      & 0.543                                           & 0.641          & 0.580                                                          \\
            LLaVA$_\text{1.0}$
                                      & \ ViT L-14 + LLaMA 2                            & caption        & 753K \ \ \ \ \         & 13.3
                                      & 0.634                                           & 0.460          & 0.528
                                      & 0.688                                           & 0.668          & 0.675
                                      & 0.610                                           & 0.511          & 0.550                                                          \\
            LLaVA$_\text{1.0}$
                                      & \ ViT L-14 + LLaMA 2                            & instruct       & 753K \ \ \ \ \         & 13.3
                                      & 0.588                                           & 0.450          & 0.505
                                      & 0.638                                           & 0.631          & 0.632
                                      & 0.615                                           & 0.541          & 0.570                                                          \\
            LLaVA$_\text{1.5}$ \cite{liu2023improvedllava}
                                      & \ ViT L-14 + Vicuna \cite{vicuna2023}           & list           & 1.2M \ \ \ \ \         & 13.4
                                      & 0.538                                           & 0.515          & 0.518
                                      & 0.591                                           & 0.783          & 0.665
                                      & 0.552                                           & 0.614          & 0.574                                                          \\
            LLaVA$_\text{1.5}$
                                      & \ ViT L-14 + Vicuna                             & caption        & 1.2M \ \ \ \ \         & 13.4
                                      & 0.632                                           & 0.453          & 0.522
                                      & 0.679                                           & 0.649          & 0.661
                                      & 0.611                                           & 0.508          & 0.549                                                          \\
            LLaVA$_\text{1.5}$
                                      & \ ViT L-14 + Vicuna                             & instruct       & 1.2M \ \ \ \ \         & 13.4
                                      & 0.572                                           & 0.498          & 0.522
                                      & 0.630                                           & 0.716          & 0.659
                                      & 0.615                                           & 0.577          & 0.582                                                          \\

            BLIP-2 \cite{li2023blip2}
                                      & \ ViT g-14 + Flant5xxl \cite{chung2022scaling}  & list           & 129M \ \ \ \ \         & 12.2
                                      & 0.544                                           & 0.557          & 0.542
                                      & 0.494                                           & 0.871          & 0.623
                                      & 0.476                                           & \ul{0.641}     & 0.538                                                          \\
            BLIP-2
                                      & \ ViT g-14 + Flant5xxl                          & caption        & 129M \ \ \ \ \         & 12.2
                                      & 0.600                                           & 0.539          & 0.561
                                      & 0.600                                           & \ul{0.893}     & 0.714
                                      & 0.523                                           & 0.626          & 0.561                                                          \\
            InstructBLIP \cite{dai2023instructblip}
                                      & \ ViT g-14 + Flant5xxl                          & list           & 129M \ \ \ \ \         & 12.3
                                      & 0.596                                           & 0.554          & 0.567
                                      & 0.613                                           & \textbf{0.897} & \ul{0.725}
                                      & 0.544                                           & 0.634          & 0.578                                                          \\
            InstructBLIP
                                      & \ ViT g-14 + Flant5xxl                          & caption        & 129M \ \ \ \ \         & 12.3
                                      & 0.639                                           & 0.487          & 0.546
                                      & 0.690                                           & 0.662          & 0.673
                                      & \ul{0.647}                                      & 0.539          & 0.581                                                          \\
            InstructBLIP
                                      & \ ViT g-14 + Flant5xxl                          & instruct       & 129M \ \ \ \ \         & 12.3
                                      & 0.529                                           & \textbf{0.604} & 0.555
                                      & 0.569                                           & 0.879          & 0.686
                                      & 0.561                                           & \textbf{0.698} & \textbf{0.615}                                                 \\
            \hline
            Ours
                                      & \ ViT L-14 + Lang$_{\text{truncated}}$          & -              & 3M \ \ \ \ \           & 1.78
                                      & \textbf{0.738}                                  & 0.530          & \textbf{0.611}
                                      & \ul{0.700}                                      & 0.712          & 0.702
                                      & 0.613                                           & 0.544          & 0.570                                                          \\
            Ours
                                      & \ ViT L-14 + Lang$_{\text{truncated}}$          & -              & 70M \ \ \ \ \          & 1.78
                                      & \ul{0.722}                                      & 0.512          & \ul{0.593}
                                      & \textbf{0.765}                                  & 0.757          & \textbf{0.758}
                                      & \textbf{0.663}                                  & 0.564          & \ul{0.603}                                                     \\
        \end{tabular}
    \end{minipage}
    \vspace{-.7em}
    \caption{
        \textbf{Comparison of different methods with top-$10$ predictions}.
        Bold numbers are the best results and underlined numbers are the second best results, same for the following tables.
    }
    \label{tab:main_results}
    \vspace{-1.em}
\end{table*}
%
\noindent
The comprehensive comparisons with other related methods,
including CLIP \cite{radford2021learning}, Open Flamingo \cite{awadalla2023openflamingo}, LLaVA \cite{liu2023visual,liu2023improvedllava}, BLIP-2 \cite{li2023blip2}, InstructBLIP \cite{dai2023instructblip}, and CaSED \cite{conti2023vocabulary}, are detailed in Table \ref{tab:main_results} with top-$10$ predictions, and Table \ref{tab:x_main_results_top5} with top-$5$ predictions.
\noindentnewline
\textbf{Preliminary}.
We construct two galleries for CLIP:
a) the base gallery, highlighted in \gray{gray}, contains reference labels only from the corresponding test dataset, e.g., CC3M validation labels for CC3M evaluation.
b) the extended gallery, includes all reference labels from the G3M training group.
\noindentnewline
Regarding CaSED \cite{conti2023vocabulary}, its performance is significantly impacted by the search gallery composition.
For a fair comparison, we evaluate CaSED using:
a) the released gallery provided with the paper, in \gray{gray}, featuring CLIP ViT-L/14 text embeddings from CC12M \cite{sharma2018conceptual};
b) the extended gallery, comprising CLIP ViT-L/14 text embeddings from COCO, SBU, CC3M, and LAION-400M, which covers our G70M training group.
CaSED can be considered a CLIP variant, with its defining aspect being the enhanced query gallery.
\noindentnewline
We evaluate other methods using their largest publicly available models.
We employ two prompt types, \textit{list} and \textit{caption}, to generate object labels from them, detailed in Section \ref{sec:x_prompt_settings}.
Also, we use the \textit{instruct} prompt for instruction-based methods, similar to its use for GPT-4V Preview \cite{gpt-4v} in \ref{sec:x_compare_with_gpt4visionpreview}.
\noindentnewline
\textbf{Analytic Comparisons}.
In the $R$ column of Table \ref{tab:main_results}, $R$ remains consistent as the number of reference labels per sample is fixed, so unaffected by prediction count.
Higher $R$ suggests top-$k$ predictions have higher semantic relevance to the reference labels.
Our method outperforms others for top-$10$ predictions across all datasets, showing our approach's ability to yield more relevant labels.
\noindentnewline
The $P$ column is sensitive to the quantity of predictions;
for instance, if we assess top-$10$ predictions but the model produces only five labels, the precision will be higher than that of the model yielding 10 predictions, according to Eq. \ref{eq:pr}.
To better understand the $P$/$R$ relationship, we plot two different precision-recall (PR) curves in Figure \ref{fig:pr_curves},
calculated by adjusting the match threshold between references and predictions, and altering $k$ for predictions.
\noindentnewline
The left column of Figure \ref{fig:pr_curves} derives from various thresholds on the similarity matrix $\bS$ in Eq. \ref{eq:cosine_similarity} with top-$10$ predictions.
The curves demonstrate a strong linear correlation due to the calculation of $P$ and $R$ from the best matches in $\bS$.
A threshold of $0.7$, for example, excludes pairs with lower similarity, reducing both $P$ and $R$ simultaneously.
The rate at which $P$ and $R$ decline with increasing thresholds reflects the overall similarity of predictions to reference labels $-$ a faster drop means the lower overall similarity.
Our method, with the gradual descent of the curves, suggests better prediction quality across all test datasets.
At a threshold of $1.0$, non-zero values of $P$ and $R$ signify that the model's predictions perfectly match the reference labels.
\noindentnewline
The right column of Figure \ref{fig:pr_curves} shows the PR curves for varying top-$k$ predictions, with the inverse correlation between $P$ and $R$, indicating their trade-off.
Our method outperforms others in both $P$ and $R$ at top-$1$ and -$3$, while at top-$5$,
Flamingo$_{\text{open}}$ and InstructBLIP saturate at the same level as top-$10$, even we double their sampling tokens for trying to generate more.
This observation demonstrates that VQA-based models are suboptimal for the task due to the lack of the ability to generate diverse labels consistently.
The plateau explain their highest $P$, but lower $R$ and $F_1$ in Table \ref{tab:main_results}.
Our method can achieve higher recall with increasing $k$, showing that it can consistently hold a $P$/$R$ balance.
\section{Ablation Studies}
\label{sec:ablation_studies}
\noindent
\textbf{Truncating the Language Decoder}.
To test our conjecture that only a subset of knowledge in LLMs is vital for the task, we reduce the decoder's size starting from LLaMA 7B.
We have found that removing intermediate transformer blocks results in a compact decoder with comparable performance.
\noindentnewline
To begin, we need to determine which transformer blocks to remove out of the 32 blocks in LLaMA 7B.
Drawing inspiration from \cite{he2022masked}, we initially fine-tuned the last third, i.e., 11 blocks, along with the final output layer.
On the other hand, motivated by the observation that the language decoder takes image embeddings as the input with a novel domain, we fine-tune the first third of the blocks, i.e., 11 blocks, and the final output layer.
This approach is premised on the hypothesis that the initial blocks might be better suited to learn the image embeddings.
As evidenced by Table \ref{tab:results_partial_training_lang_decoder}, indeed the first third of the LLaMA 7B emerges as the most significant segment.
Therefore, we decided to remove blocks after the 11$^{\text{th}}$ block.
%
\begin{table}[h]
    \tablestyle{1.8pt}{1.15}
    \begin{tabular}{z{32}|x{18.5}x{18.5}x{18.5}|x{18.5}x{18.5}x{18.5}|x{18.5}x{18.5}x{18.5}}
                                  &
        \multicolumn{3}{c|}{CC3M} &
        \multicolumn{3}{c|}{COCO} &
        \multicolumn{3}{c}{OpenImages}                                               \\
        f.t. part \ \             & R              & P              & F$_1$
                                  & R              & P              & F$_1$
                                  & R              & P              & F$_1$          \\
        \shline
        first third \ \           & \textbf{0.679} & \textbf{0.602} & \textbf{0.632}
                                  & \textbf{0.621} & \textbf{0.802} & \textbf{0.698}
                                  & \textbf{0.559} & \textbf{0.593} & \textbf{0.569} \\
        last third \ \            & 0.651          & 0.586          & 0.611
                                  & 0.585          & 0.748          & 0.654
                                  & 0.550          & 0.587          & 0.562
    \end{tabular}
    \vspace{-.7em}
    \caption{
        \textbf{Partial fine-tuning} (f.t.) results of LLaMA 7B with top-$5$ predictions, sampled by one-shot method.
        The first third encompasses the first 11 transformer blocks plus the final output layer, while the last third includes the last 11 blocks with the output layer.
    }
    \label{tab:results_partial_training_lang_decoder}
    \vspace{-1.em}
\end{table}
%
%
\begin{table}[h]
    \tablestyle{1.8pt}{1.15}
    \begin{tabular}{y{34.3}|x{18.45}x{18.45}x{18.45}|x{18.45}x{18.45}x{18.45}|x{18.45}x{18.45}x{18.45}}
                                  &
        \multicolumn{3}{c|}{CC3M} &
        \multicolumn{3}{c|}{COCO} &
        \multicolumn{3}{c}{OpenImages}                    \\
        \# params
                                  & R     & P     & F$_1$
                                  & R     & P     & F$_1$
                                  & R     & P     & F$_1$ \\
        \shline
        7.05B - 32
                                  & 0.679 & 0.602 & 0.632
                                  & 0.621 & 0.802 & 0.698
                                  & 0.559 & 0.593 & 0.569 \\
        3.00B - 11
                                  & 0.676 & 0.600 & 0.630
                                  & 0.622 & 0.805 & 0.699
                                  & 0.561 & 0.598 & 0.572 \\
        1.78B - \ 6
                                  & 0.673 & 0.598 & 0.627
                                  & 0.618 & 0.799 & 0.695
                                  & 0.560 & 0.595 & 0.570 \\
        \hline
        1.18B - \ 3
                                  & 0.670 & 0.595 & 0.624
                                  & 0.615 & 0.795 & 0.692
                                  & 0.558 & 0.593 & 0.568 \\
        0.77B - \ 1
                                  & 0.665 & 0.590 & 0.620
                                  & 0.610 & 0.790 & 0.688
                                  & 0.555 & 0.590 & 0.565
    \end{tabular}
    \vspace{-.7em}
    \caption{
        \textbf{Comparison of different language decoder sizes} with top-$5$ predictions, sampled by one-shot method.
        The number of parameters counts both the image encoder (0.43B) and the language decoder.
        It is paired with the number of transformer blocks in our language decoder, e.g., 1.78B model has 6 blocks in the decoder, denoted as 1.78B - 6.
    }
    \label{tab:results_truncaing_lang_decoder}
    \vspace{-1.em}
\end{table}
%
%
\begin{table}[h]
    \tablestyle{1.8pt}{1.15}
    \begin{tabular}{y{37}|x{18.1}x{18.1}x{18.1}|x{18.1}x{18.1}x{18.1}|x{18.1}x{18.1}x{18.1}}
        decoder w/                &
        \multicolumn{3}{c|}{CC3M} &
        \multicolumn{3}{c|}{COCO} &
        \multicolumn{3}{c}{OpenImages}                                               \\
        LLaMA
                                  & R              & P              & F$_1$
                                  & R              & P              & F$_1$
                                  & R              & P              & F$_1$          \\
        \shline
        3B \cite{touvron2023llama2}
                                  & 0.718          & 0.522          & 0.599
                                  & 0.689          & 0.702          & 0.693
                                  & 0.612          & 0.546          & 0.571          \\
        7B $\rightarrow$ 2.6B
                                  & \textbf{0.745} & \textbf{0.532} & \textbf{0.615}
                                  & \textbf{0.703} & \textbf{0.716} & \textbf{0.707}
                                  & \textbf{0.615} & \textbf{0.546} & \textbf{0.572}
    \end{tabular}
    \vspace{-.7em}
    \caption{
        \textbf{Comparison between truncated decoder and small language model} at equivalent model size with top-$10$ predictions.
    }
    \label{tab:comparison_with_llama_3b}
    \vspace{-1.em}
\end{table}
%
%
\begin{table}[t]
    \tablestyle{1.8pt}{1.15}
    \begin{tabular}{x{32}|x{18.5}x{18.5}x{18.5}|x{18.5}x{18.5}x{18.5}|x{18.5}x{18.5}x{18.5}}
                 & \multicolumn{3}{c|}{CC3M}
                 & \multicolumn{3}{c|}{COCO}
                 & \multicolumn{3}{c}{OpenImages}                                   \\
        sampling & R                              & P              & F$_1$
                 & R                              & P              & F$_1$
                 & R                              & P              & F$_1$          \\
        \shline
        greedy   & 0.661                          & \textbf{0.604} & 0.624
                 & 0.606                          & \textbf{0.802} & 0.687
                 & 0.549                          & \textbf{0.599} & 0.565          \\
        beam     & 0.641                          & 0.590          & 0.608
                 & 0.585                          & 0.772          & 0.663
                 & 0.530                          & 0.577          & 0.546          \\
        one-shot & \textbf{0.673}                 & \ul{0.598}     & \textbf{0.627}
                 & \textbf{0.618}                 & \ul{0.799}     & \textbf{0.695}
                 & \textbf{0.560}                 & \ul{0.595}     & \textbf{0.570} \\
    \end{tabular}
    \vspace{-.7em}
    \caption{
        \textbf{Comparison of different sampling methods} using top-$5$ predictions.
        The greedy and beam search sample up to 64 tokens, and takes first five generated labels as predictions.
    }
    \label{tab:results_token_sampling_methods}
    \vspace{-1.5em}
\end{table}
%
\\[0.pt]\noindent
Note that, we always retain the final output layer of LLaMA for generating the final logits.
Initially, we truncate LLaMA 7B at the 11$^{\text{th}}$ block, as illustrated in Figure \ref{fig:implementation}, resulting in a 3B model.
Table \ref{tab:results_truncaing_lang_decoder} shows that the 3B model matches the full model in performance.
To further explore the impact of the decoder size, we truncate the 3B model's decoder by removing its last 5 transformer blocks to produce a 1.78B model and find it still performs comparably to the full model.
Until the 0.77B model, which has only one transformer block, the performance has a noticeable drop but small.
\noindentnewline
The other way to construct the decoder is directly using relative small LLMs, e.g., LLaMA 3B \cite{touvron2023llama2}.
Table \ref{tab:comparison_with_llama_3b} shows our truncated decoder outperforms LLaMA 3B at the same model scale,
indicating that truncated decoders can be benefited from the better token embeddings of the larger LLMs.
Plus, truncating enables models to flexibly balance accuracy and efficiency across different model scales as in Table \ref{tab:results_truncaing_lang_decoder}.
\noindentnewline
\textbf{Sampling Strategies}.
We investigate three deterministic token sampling methods: greedy search, $3$-way beam search, and one-shot sampling.
Greedy and beam search select the highest probability token, i.e., top-$1$, at each step.
With our model, greedy and beam search suffer from the repetition issue, explained in Section \ref{sec:one-shot-sampling}.
To mitigate it for the comparison, we follow \cite{keskar2019ctrl} to penalize the logits $\x$ of the preceding generated tokens.
The sampling distribution for the next token is
\begin{aligns}
    \label{eq:penalization}
    \p = \frac{
        \exp(\x_i / (\tau \cdot \mathbbm{1}(i \in \cG)))}{
        \sum_{j} \exp(\x_j / (\tau \cdot \mathbbm{1}(j \in \cG)))
    },
\end{aligns}
where $\tau = 1.2$ is the penalization factor, $\mathbbm{1}(\cdot)$ is the indicator function, and $\cG$ is the set of preceding sampled tokens.
\noindentnewline
The results are shown in Table \ref{tab:results_token_sampling_methods}.
One-shot sampling considers label count instead of token count in greedy and beam search.
It generates more diverse labels without the \emph{repetition issue}, explaining its superior performance in $R$ and $F_1$ over greedy and beam search, though with marginally reduced $P$, consistently in top-$10$ predictions (see Table \ref{tab:x_results_token_sampling_methods_top10}).
Their top-$10$ comparisons show that, unlike one-shot sampling, increasing the number of tokens in greedy and beam search does not result in more diverse labels.
\noindentnewline
Note that our one-shot sampling could potentially encounter a \emph{competition issue}, where if multiple plausible labels share the same initial token, it would sample one of them and omit the others.
While sampling multiple times for the same token could mitigate this issue, in practice, its impact seems less critical than the repetition issue in sequential sampling.
Plus, redundant tokenization can allow multiple labels with the same starting words being returned through different token combinations.
This is tentatively indicated by our large-scale predictions in Table \ref{tab:results_ablation_on_large_scale_predictions}.
%
\begin{center}
    \includegraphics[width=1\linewidth]{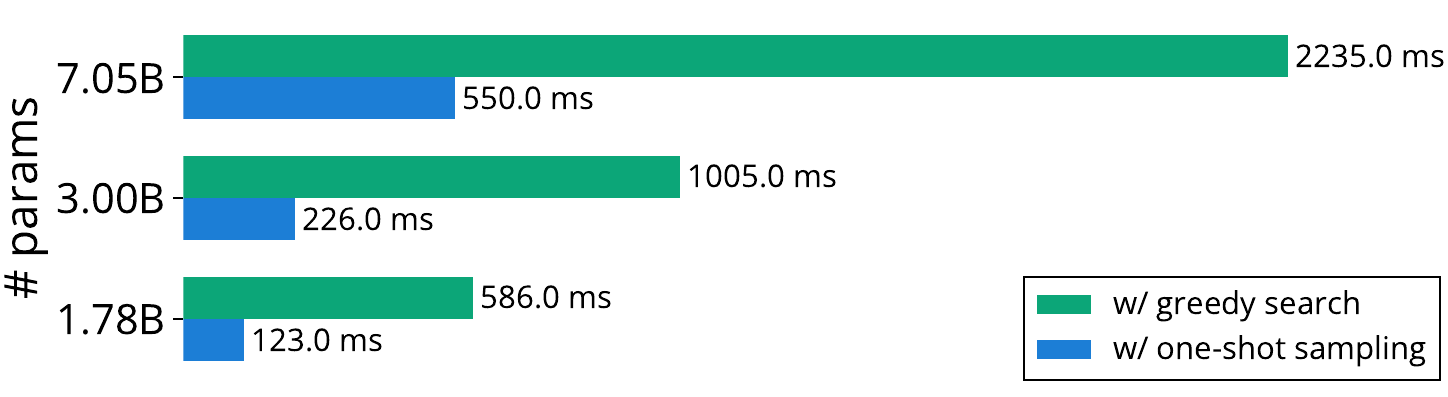}
    \vspace{-1.5em}
\end{center}
%
\textbf{Generation Efficiency}.
We combine the sampling methods with different decoder sizes to investigate their overall generation efficiency.
As illustrated above, the 1.78B model is 4.5$\times$ faster than the 7B version in inference.
Further, with one-shot sampling and truncated language model, our approach achieves 18.1$\times$ speed-up compared to the full model with greedy sampling.
The inference time is measured by the average time of generating top-$10$ labels with one-shot sampling and 64 tokens with greedy search per image.
The models run with a batch size of 1 and 16-bit Floating Point, i.e., FP16, on an A100 GPU.
Attention is without kv-cache.
\noindentnewline
\textbf{Non-causal Masking}.
In Section \ref{sec:non-causal-masking}, the non-causal masking considers two aspects:
a) prefixing image embeddings $\X_{\text{v}}$ in the input sequence, and
b) decoupling tokens from different labels to be independent.
The first ablation is to un-prefix the image embeddings as a sequential input.
Table \ref{tab:results_noncausal_modeling} shows that the prefixing is beneficial for the performance, especially with the sequential sampling strategy, i.e., greedy search.
For the one-shot sampling, the prefixing helps with a slight improvement on COCO.
\noindentnewline
The second ablation is to model tokens conditionally from different labels, also shown in Table \ref{tab:results_noncausal_modeling}.
Independent modeling is able to also provide marginal performance improvement with both greedy search and one-shot sampling, even though it provides significant gains in efficiency due to the parallelized decoding of all object labels.
%
\begin{table}[h]
    \vspace{-.5em}
    \tablestyle{1.8pt}{1.15}
    \begin{tabular}{y{32}|x{18.5}x{18.5}x{18.5}|x{18.5}x{18.5}x{18.5}|x{18.5}x{18.5}x{18.5}}
                                  &
        \multicolumn{3}{c|}{CC3M} &
        \multicolumn{3}{c|}{COCO} &
        \multicolumn{3}{c}{OpenImages}                    \\
        modeling                  & R     & P     & F$_1$
                                  & R     & P     & F$_1$
                                  & R     & P     & F$_1$ \\
        \shline
        \multicolumn{10}{l}{\textit{greedy search}}       \\
        \hline
        \ baseline                & 0.662 & 0.577 & 0.611
                                  & 0.602 & 0.754 & 0.667
                                  & 0.539 & 0.559 & 0.543 \\
        \ + prefix                & 0.664 & 0.580 & 0.613
                                  & 0.604 & 0.759 & 0.670
                                  & 0.541 & 0.563 & 0.546 \\
        \ + indep.                & 0.668 & 0.600 & 0.625
                                  & 0.609 & 0.797 & 0.688
                                  & 0.548 & 0.588 & 0.561 \\
        \hline
        \multicolumn{10}{l}{\textit{one-shot sampling}}   \\
        \hline
        baseline                  & 0.677 & 0.601 & 0.630
                                  & 0.611 & 0.790 & 0.687
                                  & 0.556 & 0.592 & 0.567
        \\
        \ + prefix                & 0.678 & 0.603 & 0.632
                                  & 0.613 & 0.792 & 0.689
                                  & 0.557 & 0.594 & 0.568 \\
        \ + indep.                & 0.679 & 0.602 & 0.632
                                  & 0.621 & 0.802 & 0.698
                                  & 0.559 & 0.593 & 0.569 \\
    \end{tabular}
    \vspace{-.7em}
    \caption{
        \textbf{Ablations for prefixing image embeddings and independent modeling of different labels} with top-$5$ predictions, generated by greedy search and one-shot sampling.
    }
    \label{tab:results_noncausal_modeling}
    \vspace{-1.em}
\end{table}
%
%
\begin{table}[h]
    \vspace{-1.em}
    \tablestyle{1.8pt}{1.15}
    \begin{tabular}{x{32}|x{18.5}x{18.5}x{18.5}|x{18.5}x{18.5}x{18.5}|x{18.5}x{18.5}x{18.5}}
                                  &
        \multicolumn{3}{c|}{CC3M} &
        \multicolumn{3}{c|}{COCO} &
        \multicolumn{3}{c}{OpenImages}                                               \\
        version                   & R              & P              & F$_1$
                                  & R              & P              & F$_1$
                                  & R              & P              & F$_1$          \\
        \shline
        \multicolumn{10}{l}{\textit{trained on G3M}}                                 \\
        \hline
        1                         & 0.673          & 0.598          & 0.627
                                  & 0.618          & 0.799          & 0.695
                                  & 0.560          & 0.595          & 0.570          \\
        2                         & \textbf{0.673} & \textbf{0.599} & \textbf{0.627}
                                  & \textbf{0.620} & \textbf{0.803} & \textbf{0.698}
                                  & \textbf{0.560} & \textbf{0.598} & \textbf{0.572} \\
        \hline
        \multicolumn{10}{l}{\textit{trained on G70M}}                                \\
        \hline
        1                         & \textbf{0.659} & \textbf{0.576} & \textbf{0.609}
                                  & \textbf{0.674} & \textbf{0.866} & \textbf{0.755}
                                  & \textbf{0.594} & \textbf{0.615} & \textbf{0.597} \\
        2                         & 0.653          & 0.572          & 0.604
                                  & 0.673          & 0.865          & 0.754
                                  & 0.593          & 0.614          & 0.596          \\
    \end{tabular}
    \vspace{-.7em}
    \caption{
        \textbf{Comparison of truncating different LLaMA versions} for the language decoder with top-$5$ predictions.
    }
    \label{tab:results_ablation_on_llama_version_top5}
    \vspace{-1.em}
\end{table}
%
%
\begin{table}[h]
    \vspace{-1.em}
    \tablestyle{1.8pt}{1.15}
    \begin{tabular}{x{32}|x{18.5}x{18.5}x{18.5}|x{18.5}x{18.5}x{18.5}|x{18.5}x{18.5}x{18.5}}
                                  &
        \multicolumn{3}{c|}{CC3M} &
        \multicolumn{3}{c|}{COCO} &
        \multicolumn{3}{c}{OpenImages}                    \\
        ranking                   & R     & P     & F$_1$
                                  & R     & P     & F$_1$
                                  & R     & P     & F$_1$ \\
        \shline
        -                         & 0.673 & 0.598 & 0.627
                                  & 0.618 & 0.799 & 0.695
                                  & 0.560 & 0.595 & 0.570 \\
        full                      & 0.673 & 0.598 & 0.627
                                  & 0.619 & 0.800 & 0.695
                                  & 0.562 & 0.597 & 0.572
    \end{tabular}
    \vspace{-.7em}
    \caption{
        \textbf{Comparison of different strategies for ranking top-$5$ predictions}.
        The first row ranks predictions using initial token probabilities, whereas the second row uses full label probabilities, derived by multiplying token probabilities.
    }
    \label{tab:results_ablation_ranking_one_shot}
    \vspace{-1.em}
\end{table}
%
%
\begin{table}[h]
    \vspace{-1.em}
    \tablestyle{1.8pt}{1.15}
    \begin{tabular}{y{32}|x{18.5}x{18.5}x{18.5}|x{18.5}x{18.5}x{18.5}|x{18.5}x{18.5}x{18.5}}
                                  &
        \multicolumn{3}{c|}{CC3M} &
        \multicolumn{3}{c|}{COCO} &
        \multicolumn{3}{c}{OpenImages}                                               \\
        \ method                  & R              & P              & F$_1$
                                  & R              & P              & F$_1$
                                  & R              & P              & F$_1$          \\
        \shline
        \ \gray{CLIP}
                                  & \gray{0.752}   & \gray{0.360}   & \gray{0.483}
                                  & \gray{0.715}   & \gray{0.430}   & \gray{0.536}
                                  & \gray{0.666}   & \gray{0.387}   & \gray{0.485}   \\
        \ CLIP
                                  & 0.615          & 0.332          & 0.427
                                  & 0.576          & 0.411          & 0.478
                                  & 0.506          & 0.334          & 0.399          \\
        \hline
        \ ours                    & \textbf{0.868} & \textbf{0.394} & \textbf{0.538}
                                  & \textbf{0.930} & \textbf{0.499} & \textbf{0.649}
                                  & \textbf{0.874} & \textbf{0.448} & \textbf{0.589} \\
    \end{tabular}
    \vspace{-.7em}
    \caption{
        \textbf{Large-scale top-$100$ predictions} with the same settings in Table \ref{tab:main_results}.
    }
    \label{tab:results_ablation_on_large_scale_predictions}
    \vspace{-1.em}
\end{table}
%
\noindentnewline
\textbf{Different LLaMA Versions}.
In Table \ref{tab:results_ablation_on_llama_version_top5}, we compare two truncated versions of LLaMA, namely 1.78B models of LLaMA 1 \cite{touvron2023llama} and LLaMA 2 \cite{touvron2023llama2}.
LLaMA 2 marginally outperforms LLaMA 1 trained on G3M, and has comparable results trained on G70M.
\noindentnewline
\textbf{Ranking Predictions}.
Our one-shot sampling method selects the final top-$k$ labels based on the probabilities of their initial tokens.
Table \ref{tab:results_ablation_ranking_one_shot} demonstrates the effectiveness of this approach compared to using full label probabilities.
Further details on ranking strategies can be found in \ref{sec:x_results_of_ranking_predictions}.
\noindentnewline
\textbf{Large-scale Prediction}.
We evaluate our method on large-scale prediction, i.e., top-$100$ predictions, with the same settings as in Table \ref{tab:main_results}.
Table \ref{tab:results_ablation_on_large_scale_predictions} shows our method's consistent ability to predict diverse labels as the number of predictions increases, where $R$ and $F_1$ are improved, and $P$ is decreased.
Besides, CLIP \cite{radford2021learning} has a similar trend, but its performance is much lower than ours.
Further, with inflating its gallery from \gray{base} to the extended one, CLIP has a performance drop across all datasets, also observed in \cite{conti2023vocabulary}.
%
\begin{figure}[h]
    \vspace{-1.em}
    \centering
    \includegraphics[width=1\linewidth]{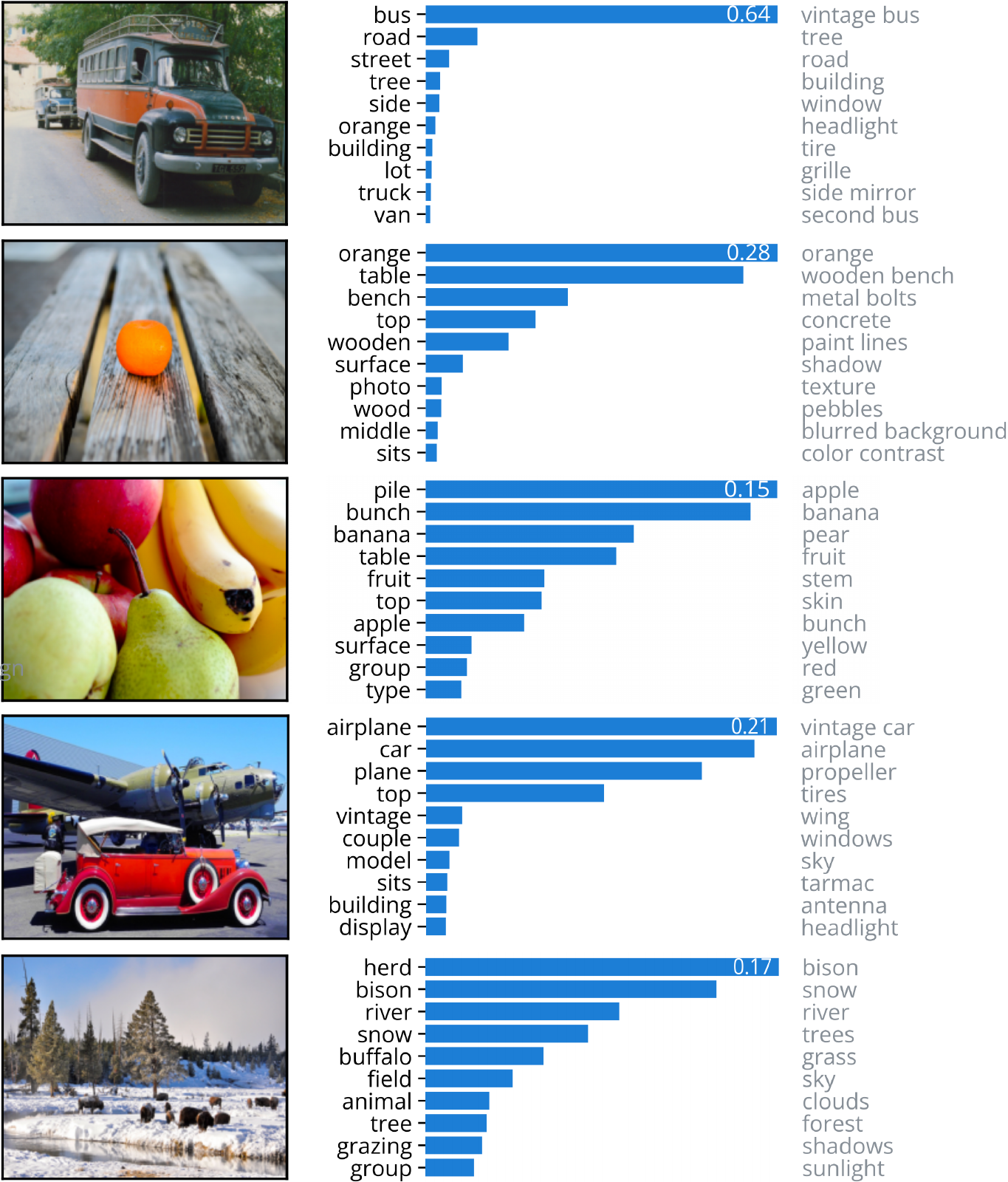}
    \vspace{-1.5em}
    \caption{
        \textbf{Qualitative results} with top-$10$ predictions.
        The top bar is with the first prediction's probability.
        The right \textcolor{mygray}{gray} column displays GPT-4V Preview \cite{gpt-4v}'s predictions.
        For extensive results of $336$ images, refer to Section \ref{sec:x_results_visualization}.
    }
    \label{fig:results}
    \vspace{-1.5em}
\end{figure}
%
\section{Conclusion}
\label{sec:conclusion}
\noindent
We have presented an auto-regressive framework for object recognition based on next token prediction, efficiently generating labels with one-shot sampling in parallel and intuitively depending only on the number of required labels.

%% file: sec/X_suppl.tex
\clearpage
\setcounter{figure}{0}
\setcounter{table}{0}
\setcounter{equation}{0}
\setcounter{subsection}{0}
\renewcommand{\thesubsection}{A.\arabic{subsection}}
\renewcommand{\thefigure}{A.\arabic{figure}}
\renewcommand{\thetable}{A.\arabic{table}}
\renewcommand{\theequation}{A.\arabic{equation}}
%
%
%
\noindent
\section*{A. Appendix}
\subsection{Compare with GPT-4V Preview}
\label{sec:x_compare_with_gpt4visionpreview}
\begin{figure}[t]
    \vspace{.5em}
    \centering
    \begin{minipage}{0.5\linewidth}
        \includegraphics[width=\linewidth]{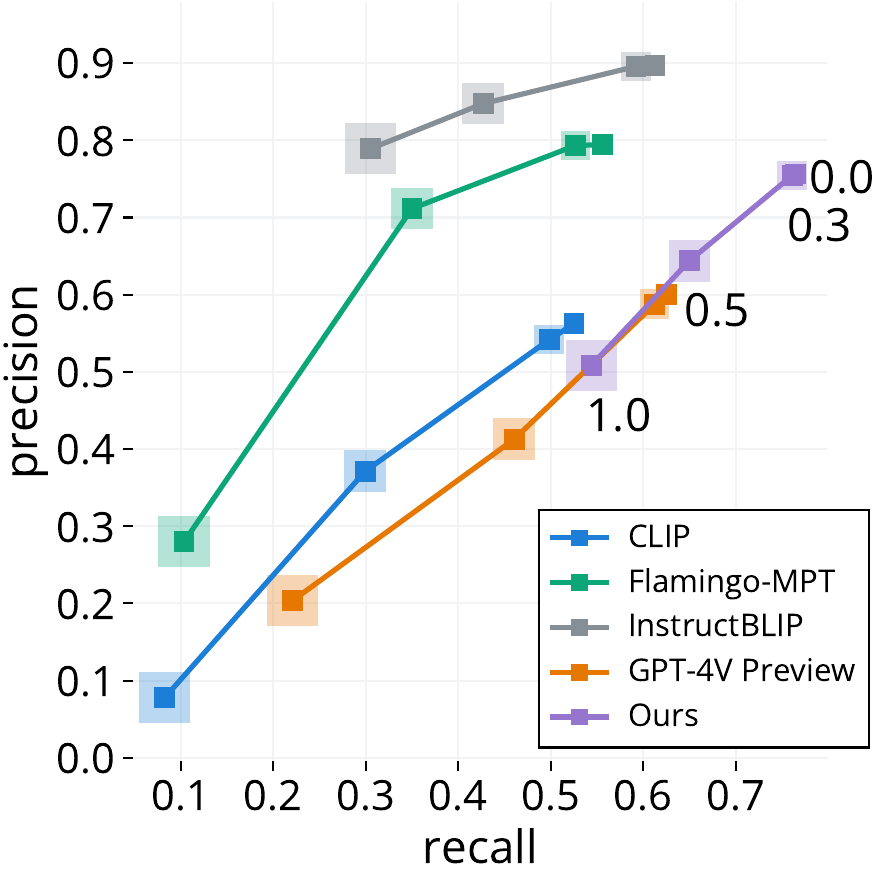}
    \end{minipage}%
    \begin{minipage}{0.5\linewidth}
        \includegraphics[width=\linewidth]{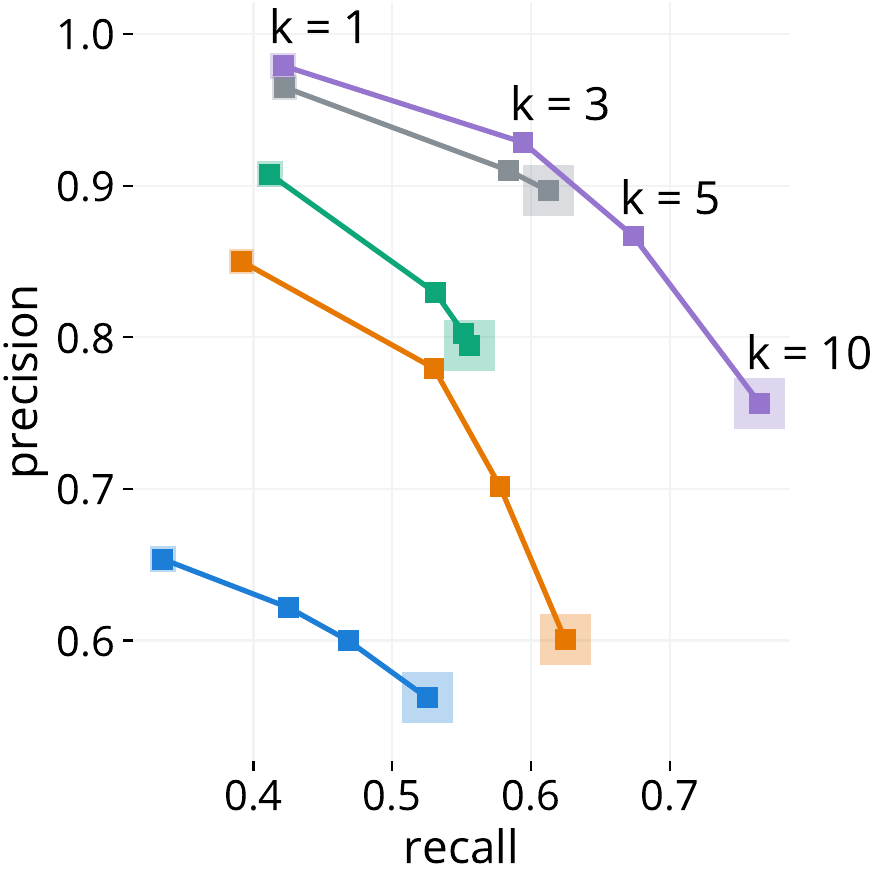}
    \end{minipage}%
    \vspace{-.7em}
    \caption{
        \textbf{Precision-recall (PR) curves} on COCO validation subset.
        The same settings as in Figure \ref{fig:pr_curves}.
    }
    \label{fig:x_pr_curves_coco_w_gpt4v}
\end{figure}

Since the GPT-4V(ision) Preview \cite{gpt-4v} is also able to generate object labels for images, we compare our method with it for the recognition task.
The API parameters for the GPT-4V Preview \cite{gpt-4v} are: input image size is $256^2$, temperature is zero for deterministic predictions, and detail is low with sampling 65 output tokens.
The model version from API is {\mytexttt{gpt-4-1106-vision-preview}}.
We prompt it to generate ten main object labels as its top-$10$ predictions with the following instruction:
%
{\begin{center}
    \tablestyle{1.8pt}{1.15}
    \begin{tabular}{y{233}}
        the instruction for OpenAI GPT-4-vision-preview API\footnotemark                                            \\
        \shline
        Describe every detail in the image by listing ten main object labels.
        The answer should only contain the object labels separated by a comma, for example, ``car, airplane, dog''. \\
        \hline
    \end{tabular}
    \footnotetext{
        \href{https://platform.openai.com/docs/guides/vision}{\textcolor{black}{platform.openai.com/docs/guides/vision}}.
    }
\end{center}}
%
\noindent
Due to the API request limit, we are able to evaluate it on a subset of the COCO validation split, which contains 4359 out of 5000 images in total.
We compare various methods in Table \ref{tab:x_results_top10_coco_w_gpt4v} with top-$10$ predictions, showing that our method performs better than the GPT-4V Preview \cite{gpt-4v} across all metrics, and the GPT-4V Preview has the second-highest $R$.
The PR-curves are illustrated in Figure \ref{fig:x_pr_curves_coco_w_gpt4v}, indicating that our method has a better $P$/$R$ trade-off.
Since GPT-4V Preview consistently generates ten labels for each image, its $P$ is also low compared to Flamingo$_{\text{open}}$ and InstructBLIP.
%
\begin{table}[h]
    \tablestyle{1.8pt}{1.15}
    \begin{tabular}{y{106}|x{24}|x{18.5}x{18.5}x{18.5}}
         &                & \multicolumn{3}{c}{COCO}                  \\
        \ method
         & prompt
         & R              & P                        & F$_1$          \\
        \shline
        \ CLIP \cite{radford2021learning}
         & -
         & 0.525          & 0.562                    & 0.540          \\
        \ Flamingo$_{\text{open}}$ \cite{awadalla2023openflamingo} w/ MPT \cite{MosaicML2023Introducing}
         & list
         & 0.556          & \ul{0.794}               & 0.647          \\
        \ InstructBLIP \cite{dai2023instructblip}
         & list
         & 0.613          & \textbf{0.897}           & \ul{0.725}     \\
        \ GPT-4V Preview \cite{gpt-4v}
         & instruct
         & \ul{0.625}     & 0.601                    & 0.610          \\
        \hline
        \ Ours
         & -
         & \textbf{0.765} & 0.756                    & \textbf{0.758}
    \end{tabular}
    \vspace{-.7em}
    \caption{
        \textbf{Comparison with top-$10$ predictions} on COCO validation subset.
    }
    \label{tab:x_results_top10_coco_w_gpt4v}
    \vspace{-1.em}
\end{table}
%
\noindentnewline
\textbf{Cross-Validation}.
As we mentioned in Section \ref{sec:non-causal-masking}, the reference labels extracted from the raw captions are imperfect and incomplete.
To verify that our method generalizes well to predict plausible labels, we conduct a cross-validation on the COCO validation subset, treating the GPT-4V Preview's predictions as reference labels to evaluate others.
Table \ref{tab:x_results_top10_coco_w_gpt4v_as_reference} demonstrates that our method consistently matches the performance across all metrics as presented in Table \ref{tab:main_results}, in which our method ranks first in $R$ and $F_1$.
Again, the lower $P$ for our method is due to the fact that our model predicts the required number of labels, while others with a higher $P$ presumably predict less than ten labels.
Regarding $R$, LLaVA$_\text{1.0}$ \cite{liu2023visual} ranks second in performance.
%
%
%
\begin{table}[h]
    \vspace{-.5em}
    \tablestyle{1.8pt}{1.15}
    \begin{tabular}{y{106}|x{24}|x{18.5}x{18.5}x{18.5}}
         &                & \multicolumn{3}{c}{COCO}                  \\
        \ method
         & prompt
         & R              & P                        & F$_1$          \\
        \shline
        \ CLIP \cite{radford2021learning}
         & -
         & 0.467          & 0.509                    & 0.485          \\
        \ CaSED \cite{conti2023vocabulary}
         & -
         & 0.535          & 0.562                    & 0.546          \\
        \ Flamingo$_{\text{open}}$ \cite{awadalla2023openflamingo} w/ MPT \cite{MosaicML2023Introducing}
         & list
         & 0.517          & \textbf{0.760}           & \ul{0.609}     \\
        \ LLaVA$_\text{1.0}$ \cite{liu2023visual}
         & caption
         & \ul{0.593}     & 0.599                    & 0.595          \\
        \ LLaVA$_\text{1.5}$ \cite{liu2023improvedllava}
         & caption
         & 0.576          & 0.572                    & 0.573          \\
        \ BLIP-2 \cite{li2023blip2}
         & caption
         & 0.498          & \ul{0.736}               & 0.590          \\
        \ InstructBLIP \cite{dai2023instructblip}
         & list
         & 0.505          & 0.731                    & 0.594          \\
        \ GPT-4V Preview \cite{gpt-4v}
         & instruct
         & 1.000          & 1.000                    & 1.000          \\
        \hline
        \ Ours
         & -
         & \textbf{0.632} & 0.651                    & \textbf{0.641} \\
        \ \gray{Ours w/ top-$100$}
         & -
         & \gray{0.823}   & \gray{0.473}             & \gray{0.600}
    \end{tabular}
    \vspace{-.7em}
    \caption{
        \textbf{Comparison with top-$10$ predictions} on COCO validation subset, viewing GPT-4V Preview's predictions as reference labels.
        \gray{Gray row} shows our top-$100$ predictions.
    }
    \label{tab:x_results_top10_coco_w_gpt4v_as_reference}
    \vspace{-1.em}
\end{table}
%
\subsection{Ranking Predictions}
\label{sec:x_results_of_ranking_predictions}
\noindent
We ablate ranking strategies for the predictions produced by our model.
Given an image, our model generates $K$ labels $\cL = \{L_1, \dots, L_K\}$.
Each label $L_k$ has $T_k + 1$ tokens, including the special token [SEP] for the delimiter.
\noindentnewline
\textbf{Ranking by CLIP Score}.
The first strategy is to rank the predictions by the CLIP score:
\begin{aligns}
    \text{clip}(L_k) = f_{\text{CLIP}}(\text{image}, \text{label }L_k),
\end{aligns}
where $f_{\text{CLIP}}$ is the CLIP model \cite{radford2021learning} with the image encoder of ViT-L/14 and the language encoder.
The CLIP score is based on cosine distance in the embedding space.
\noindentnewline
\textbf{Ranking by Probability}.
The second strategy is to rank the predictions by their probabilities in Eq. \ref{eq:multi-label}:
\begin{aligns}
    \text{prob}(L_k) = \prod_{t=1}^{T_k + 1} P(\w^k_t | \w^k_{<t}, \X),
\end{aligns}
in which the probability of each label is the product of the individual probabilities of its tokens, including the delimiter token [SEP].
If greedy and beam search sample a particular label multiple times, we sum up the probabilities as its final probability.
\noindentnewline
\textbf{Ranking by Perplexity}.
The third one is to rank the predictions by their perplexities.
The perplexity is computed with the fixed length $T_k + 1$ for each label:
\begin{aligns}
    \text{ppl}(L_k) = \exp \left[ -\frac{1}{T_k + 1}\sum_{t=1}^{T_k + 1} \log P(\w^k_t | \w^k_{<t}, \X) \right].
\end{aligns}
If the greedy and beam search sample a particular label multiple times, we use its minimum perplexity to ensure optimal selection and accuracy.
\noindentnewline
\textbf{Ranking by Cross-Modal Similarity Score}.
The last one is to rank predictions by their cross-modal similarity scores, computed with the image and label token embeddings:
\begin{aligns}
    \text{sim}(L_k) = \frac{1}{T_k} \sum_{t=1}^{T_k} d(\w^k_t, \X_\text{v}),
\end{aligns}
where $d$ is the euclidean distance averaged over all the image token embeddings for each label token embedding $\w_t^k$:
\begin{aligns}
    d(\w^k_t, \X_\text{v}) = \frac{1}{M} \sum_{i=1}^{M} \sqrt{2 - 2 \cdot \frac{\w^k_t \cdot \x^{\text{v}}_i}{\left\| \w^k_t \right\|_2 \cdot \left\| \x^{\text{v}}_i \right\|_2}},
\end{aligns}
where $M$ is the number of image tokens.
This similarity is also called compatibility score to measure the compatibility between image and label embeddings, which motivates us to select the predictions that are compatible with the corresponding images.
In other words, the closer the label token embeddings are to the image token embeddings, the more likely the label is the correct prediction.
\noindentnewline
\textbf{Results}.
Table \ref{tab:x_results_ranking_strategies} compares the above four ranking strategies using top-$5$ predictions across different sampling methods for our 1.78B model trained on G3M.
The greedy and 3-way beam search samples 64 tokens for each image.
Since one-shot sampling yields ordered predictions, we sample 10 labels per image and utilize ranking strategies to select the final top-$5$ predictions.
%
\begin{table}[t]
    \tablestyle{1.8pt}{1.15}
    \begin{tabular}{x{32}|x{18.5}x{18.5}x{18.5}|x{18.5}x{18.5}x{18.5}|x{18.5}x{18.5}x{18.5}}
                                    &
        \multicolumn{3}{c|}{greedy} &
        \multicolumn{3}{c|}{beam}   &
        \multicolumn{3}{c}{one-shot}                                                            \\
        ranking                     &
        R                           & P                 & F$_{1}$           &
        R                           & P                 & F$_{1}$           &
        R                           & P                 & F$_{1}$                               \\
        \shline
        \multicolumn{10}{l}{\textit{CC3M}}                                                      \\
        \hline
        -                           & \ul{0.661}        & \textbf{0.604}    & \ul{0.624}
                                    & 0.641             & 0.590             & 0.608
                                    & 0.673             & 0.598             & 0.627             \\
        clip                        & 0.646             & \ul{0.604}        & 0.617
                                    & 0.630             & 0.594             & 0.605
                                    & 0.643             & 0.588             & 0.608             \\
        prob                        & 0.659             & 0.602             & 0.622
                                    & -                 & -                 & -
                                    & \textbf{0.673}    & 0.598             & \textbf{0.627}    \\
        ppl                         & 0.614             & 0.563             & 0.581
                                    & -                 & -                 & -
                                    & 0.509             & 0.466             & 0.484             \\
        sim                         & 0.611             & 0.564             & 0.581
                                    & 0.598             & 0.557             & 0.571
                                    & 0.594             & 0.531             & 0.556             \\
        \hline
        \multicolumn{10}{l}{\textit{COCO}}                                                      \\
        \hline
        -                           & \ul{0.606}        & \textbf{0.802}    & 0.687
                                    & 0.585             & 0.772             & 0.663
                                    & 0.618             & 0.799             & 0.695             \\
        clip                        & 0.590             & 0.792             & 0.673
                                    & 0.573             & 0.772             & 0.654
                                    & 0.592             & 0.773             & 0.668             \\
        prob                        & 0.603             & 0.796             & \underline{0.683}
                                    & -                 & -                 & -
                                    & \textbf{0.619}    & \ul{0.800}        & \textbf{0.695}    \\
        ppl                         & 0.578             & 0.748             & 0.649
                                    & -                 & -                 & -
                                    & 0.528             & 0.640             & 0.577             \\
        sim                         & 0.576             & 0.747             & 0.647
                                    & 0.552             & 0.724             & 0.623
                                    & 0.576             & 0.717             & 0.637             \\
        \hline
        \multicolumn{10}{l}{\textit{OpenImages}}                                                \\
        \hline
        -                           & 0.549             & 0.599             & 0.565
                                    & 0.530             & 0.577             & 0.546
                                    & 0.560             & 0.595             & \ul{0.570}        \\
        clip                        & 0.540             & \textbf{0.598}    & 0.560
                                    & 0.525             & 0.580             & 0.544
                                    & 0.543             & 0.591             & 0.559             \\
        prob                        & \textbf{0.580}    & 0.576             & 0.569
                                    & -                 & -                 & -
                                    & 0.562             & \underline{0.597} & \textbf{0.572}    \\
        ppl                         & \underline{0.577} & 0.571             & 0.565
                                    & -                 & -                 & -
                                    & 0.495             & 0.505             & 0.496             \\
        sim                         & 0.575             & 0.571             & 0.564
                                    & 0.509             & 0.553             & 0.524
                                    & 0.527             & 0.547             & 0.532             \\
    \end{tabular}
    \vspace{-.7em}
    \caption{
        \textbf{Comparison of different ranking strategies for various sampling methods} with top-$5$ predictions.
        In the case of ``-'', no ranking strategy is used, and one-shot sampling directly outputs the top-$5$ labels.
    }
    \label{tab:x_results_ranking_strategies}
    \vspace{-1.em}
\end{table}
%
%
\begin{table}[h]
    \tablestyle{1.8pt}{1.15}
    \begin{tabular}{x{32}|x{18.5}x{18.5}x{18.5}|x{18.5}x{18.5}x{18.5}|x{18.5}x{18.5}x{18.5}}
                & \multicolumn{3}{c|}{CC3M}
                & \multicolumn{3}{c|}{COCO}
                & \multicolumn{3}{c}{OpenImages}                 \\
        ranking & R                              & P     & F$_1$
                & R                              & P     & F$_1$
                & R                              & P     & F$_1$ \\
        \shline
        -       & 0.545                          & 0.568 & 0.549
                & 0.548                          & 0.794 & 0.643
                & 0.526                          & 0.655 & 0.576 \\
        clip    & 0.551                          & 0.574 & 0.555
                & 0.552                          & 0.801 & 0.648
                & 0.527                          & 0.657 & 0.577 \\
    \end{tabular}
    \vspace{-.7em}
    \caption{
        \textbf{Comparison of different ranking strategies} with top-$5$ predictions for Flamingo$_\text{open}$ + MPT.
    }
    \label{tab:x_results_flamingo_without_ranking}
    \vspace{-1.em}
\end{table}
%
\noindentnewline
The overall best ranking strategy is using probability for greedy search and one-shot sampling, and using CLIP score for beam search.
For $R$, one-shot sampling with probability ranks first on CC3M and COCO, and the greedy search with probability leads on OpenImages.
The greedy search with probability has a slightly higher $P$ than one-shot sampling with probability, but the latter has a better overall $F_1$.
\noindentnewline
For greedy search, the compatibility score has the same performance as the perplexity.
For one-shot sampling, the compatibility score is better than the perplexity.
Without a ranking strategy, one-shot sampling matches the performance of probability-based ranking, showing its effectiveness in using top-$k$ initial tokens to decide the final top-$k$ predictions.
\noindentnewline
No ranking strategy outperforms the CLIP score for both greedy and beam search, yet we apply CLIP score to other models like Flamingo, BLIP-2, InstructBLIP, and LLaVA.
For BLIP-2, InstructBLIP, and LLaVA, whose outputs are sentences, the CLIP score is the only choice for ranking.
But for Flamingo, since it has a same format as ours, we can test its performance without ranking strategy.
Because it saturates at top-$10$, we only report its top-$5$ comparison.
The results are shown in Table \ref{tab:x_results_flamingo_without_ranking}, showing that the CLIP score is the optimal ranking strategy for those models.
\subsection{Additional Results}
\label{sec:x_additional_results}
\begin{figure}[h]
    \centering
    \begin{minipage}{0.5\linewidth}
        \includegraphics[width=\linewidth]{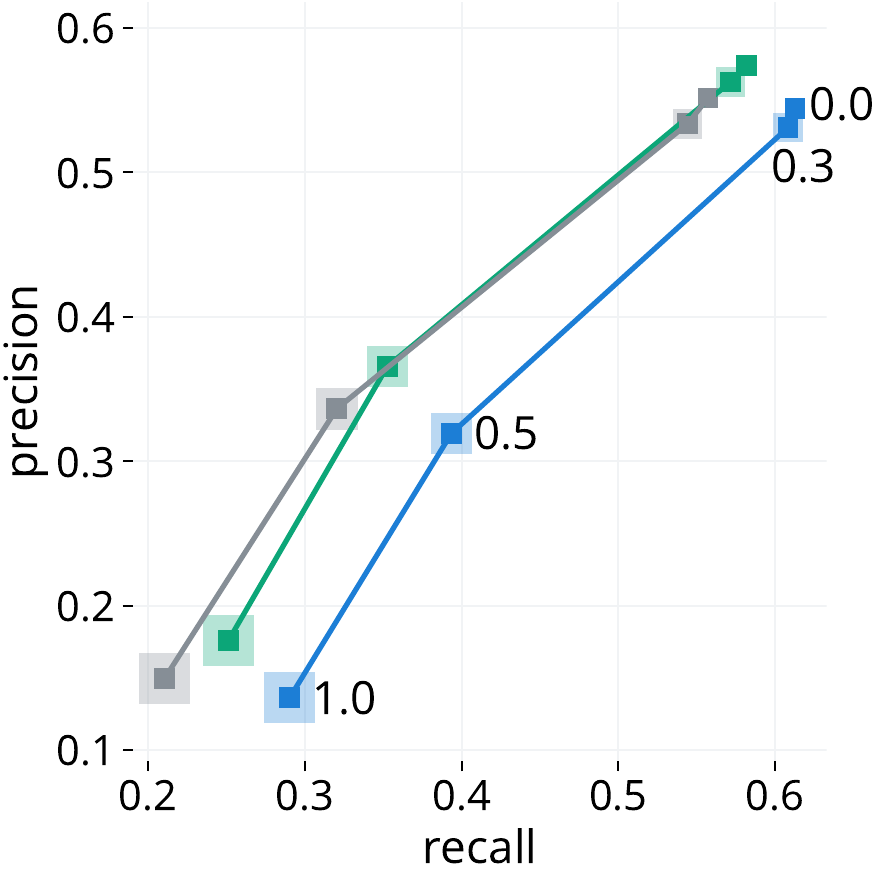}
    \end{minipage}%
    \begin{minipage}{0.5\linewidth}
        \includegraphics[width=\linewidth]{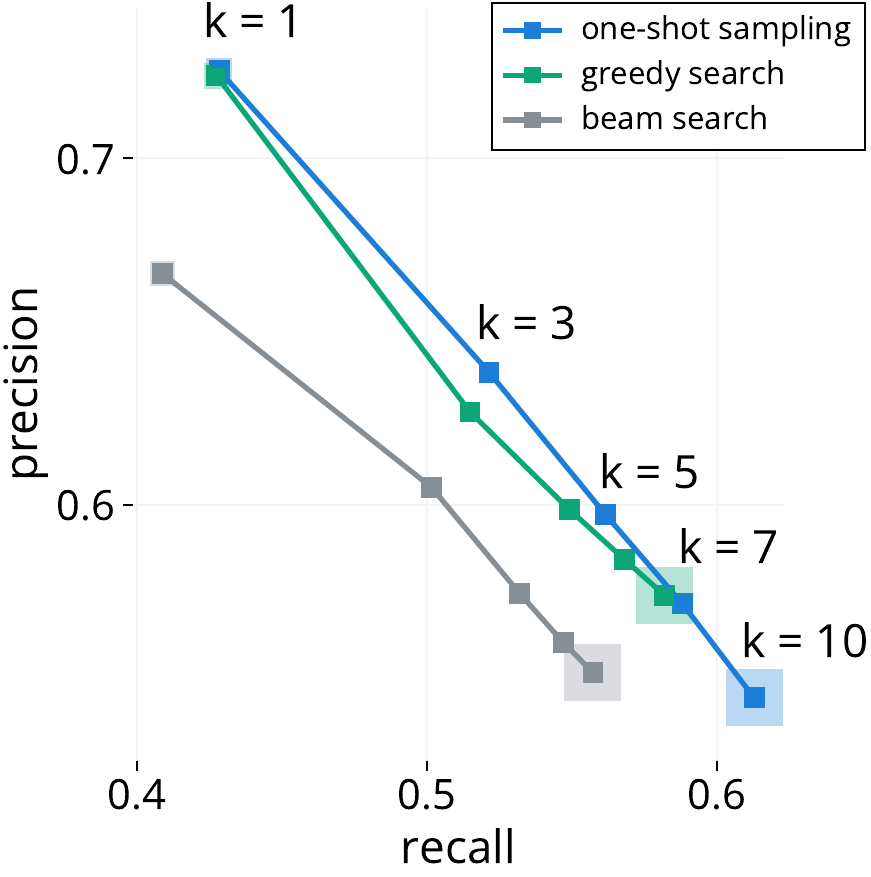}
    \end{minipage}%
    \vspace{-.7em}
    \caption{
        \textbf{Precision-recall (PR) curves of different sampling methods} on OpenImages validation split with top-$10$ predictions.
        The same settings as in Figure \ref{fig:pr_curves}.
    }
    \label{fig:x_pr_curves_sampling_methods}
    \vspace{-1.em}
\end{figure}

\noindent
In this section, we present additional results, mainly with top-$10$ predictions, for ablation studies.
\noindentnewline
\textbf{Ablation on Truncating the Decoder}.
We compare the results of different truncating sizes of the language decoder with top-$10$ predictions in Table \ref{tab:x_results_truncaing_lang_decoder_top10}.
There is a small performance drop, $0.745 \rightarrow 0.738$ in $R$ on CC3M, with truncating the decoder from 3B to 1.78B, while the performances on COCO and OpenImages remain the same.
\noindentnewline
\textbf{Ablation on Sampling Methods}.
We compare sampling methods, i.e., greedy search, 3-way beam search, and one-shot sampling, with top-$10$ predictions in Table \ref{tab:x_results_token_sampling_methods_top10}.
The results, consistent with those in Table \ref{tab:results_token_sampling_methods}, indicate that one-shot sampling surpasses greedy and beam search in $R$ and $F_1$ scores but falls short in $P$ when considering top-$10$ predictions.
The reason is that greedy and beam search produce {\small{$\sim$}}$7$ labels average per image in top-$10$ due to the repetition issue.
Figure \ref{fig:x_pr_curves_sampling_methods} (right side) demonstrates saturation around $k = 7$, accounting for their higher $P$ in top-10 predictions.
This ablation study shows that greedy and beam search do not produce more diverse predictions with increasing number of tokens.
%
%
\noindentnewline
\textbf{Ablation on LLaMA Versions}.
Table \ref{tab:x_results_ablation_on_llama_version_top10} compares the results of different LLaMA versions for the language decoder with top-$10$ predictions.
The top-10 results are consistent with Table \ref{tab:results_ablation_on_llama_version_top5}, showing LLaMA 2 is slightly better than LLaMA 1 on G3M, and comparable on G70M.
\noindentnewline
\textbf{Ablation on Embedding Models in Evaluation Metric}.
The evaluation metric is based on embedding models to compute the similarity $\bS_{ij}$ in Eq. \ref{eq:cosine_similarity}.
To verify the robustness of our method, we compare the results using CLIP ViT-L/14 \cite{radford2021learning} as the metric embedding model in Table \ref{tab:results_ablation_on_embedding_models_in_evaluation}.
Our results are from the 1.78B model trained on G70M, and the others are from the best settings in Table \ref{tab:main_results}.
Our method consistently outperforms others in $R$ and $F_1$ scores, and is competitive in $P$.
\noindentnewline
\textbf{Ablation on Training Epochs}.
We conduct an ablation study on training epochs for our 1.78B model on G3M.
Table \ref{tab:x_ablation_on_training_epochs} shows the results with top-$10$ predictions, indicating that training more epochs improves the performance.
\noindentnewline
\textbf{Additional Main Results}.
Table \ref{tab:x_main_results_top5} shows the main results with top-$5$ predictions, consistent with those in Table \ref{tab:main_results}.
The performance drop on CC3M for models trained on G3M versus G70M stems from a data distribution shift.
%
\begin{table}[h]
    \tablestyle{1.8pt}{1.15}
    \begin{tabular}{y{34.3}|x{18.45}x{18.45}x{18.45}|x{18.45}x{18.45}x{18.45}|x{18.45}x{18.45}x{18.45}}
                                  &
        \multicolumn{3}{c|}{CC3M} &
        \multicolumn{3}{c|}{COCO} &
        \multicolumn{3}{c}{OpenImages}                    \\
        \# params                 & R     & P     & F$_1$
                                  & R     & P     & F$_1$
                                  & R     & P     & F$_1$ \\
        \shline
        7.05B - 32                & 0.748 & 0.534 & 0.617
                                  & 0.699 & 0.710 & 0.702
                                  & 0.613 & 0.543 & 0.569 \\
        3.00B - 11                & 0.745 & 0.532 & 0.615
                                  & 0.703 & 0.716 & 0.707
                                  & 0.615 & 0.546 & 0.572 \\
        1.78B - \ 6               & 0.738 & 0.530 & 0.611
                                  & 0.698 & 0.712 & 0.702
                                  & 0.613 & 0.544 & 0.570 \\
        \hline
        1.18B - \ 3               & 0.736 & 0.530 & 0.611
                                  & 0.697 & 0.713 & 0.703
                                  & 0.612 & 0.547 & 0.571 \\
        0.77B - \ 1               & 0.731 & 0.529 & 0.608
                                  & 0.693 & 0.708 & 0.698
                                  & 0.609 & 0.547 & 0.569 \\
    \end{tabular}
    \vspace{-.7em}
    \caption{
        \textbf{Comparison of different language decoder sizes} with top-$10$ predictions.
        The same settings as in Table \ref{tab:results_truncaing_lang_decoder}.
    }
    \label{tab:x_results_truncaing_lang_decoder_top10}
    \vspace{-1.em}
\end{table}
%
%
\begin{table}[h]
    \tablestyle{1.8pt}{1.15}
    \begin{tabular}{x{32}|x{18.5}x{18.5}x{18.5}|x{18.5}x{18.5}x{18.5}|x{18.5}x{18.5}x{18.5}}
                 & \multicolumn{3}{c|}{CC3M}
                 & \multicolumn{3}{c|}{COCO}
                 & \multicolumn{3}{c}{OpenImages}                                   \\
        sampling & R                              & P              & F$_1$
                 & R                              & P              & F$_1$
                 & R                              & P              & F$_1$          \\
        \shline
        greedy   & \ul{0.708}                     & \textbf{0.568} & \textbf{0.621}
                 & \ul{0.655}                     & \textbf{0.755} & 0.696
                 & \ul{0.582}                     & \textbf{0.574} & \ul{0.569}     \\
        beam     & 0.681                          & \ul{0.557}     & 0.604
                 & 0.623                          & \ul{0.725}     & 0.665
                 & 0.557                          & \ul{0.552}     & 0.546          \\
        one-shot & \textbf{0.738}                 & 0.530          & \ul{0.611}
                 & \textbf{0.698}                 & 0.712          & \textbf{0.702}
                 & \textbf{0.613}                 & 0.544          & \textbf{0.570} \\
    \end{tabular}
    \vspace{-.7em}
    \caption{
        \textbf{Comparison of different sampling methods} with top-$10$ predictions.
        The greedy and beam search sample 128 tokens for each image without ranking strategies.
    }
    \label{tab:x_results_token_sampling_methods_top10}
    \vspace{-1.em}
\end{table}
%
%
\begin{table}[h]
    \tablestyle{1.8pt}{1.15}
    \begin{tabular}{x{32}|x{18.5}x{18.5}x{18.5}|x{18.5}x{18.5}x{18.5}|x{18.5}x{18.5}x{18.5}}
                                  &
        \multicolumn{3}{c|}{CC3M} &
        \multicolumn{3}{c|}{COCO} &
        \multicolumn{3}{c}{OpenImages}                                               \\
        version                   & R              & P              & F$_1$
                                  & R              & P              & F$_1$
                                  & R              & P              & F$_1$          \\
        \shline
        \multicolumn{10}{l}{\textit{trained on G3M}}                                 \\
        \hline
        1                         & 0.738          & 0.530          & 0.611
                                  & 0.698          & 0.712          & 0.702
                                  & 0.613          & 0.544          & 0.570          \\
        2                         & \textbf{0.740} & \textbf{0.531} & \textbf{0.612}
                                  & \textbf{0.700} & \textbf{0.714} & \textbf{0.705}
                                  & \textbf{0.614} & \textbf{0.547} & \textbf{0.571} \\
        \hline
        \multicolumn{10}{l}{\textit{trained on G70M}}                                \\
        \hline
        1                         & \textbf{0.722} & \textbf{0.512} & \textbf{0.593}
                                  & \textbf{0.765} & \textbf{0.757} & \textbf{0.758}
                                  & \textbf{0.663} & \textbf{0.564} & \textbf{0.603} \\
        2                         & 0.721          & 0.512          & 0.593
                                  & 0.765          & 0.756          & 0.758
                                  & 0.662          & 0.563          & 0.602          \\
    \end{tabular}
    \vspace{-.7em}
    \caption{
        \textbf{Comparison of truncating different LLaMA versions} for the language decoder with top-$10$ predictions.
    }
    \label{tab:x_results_ablation_on_llama_version_top10}
    \vspace{-1.em}
\end{table}
%
%
\begin{table}[h]
    \tablestyle{1.8pt}{1.15}
    \begin{tabular}{y{33}|x{18.4}x{18.4}x{18.4}|x{18.4}x{18.4}x{18.4}|x{18.4}x{18.4}x{18.4}}
                                  & \multicolumn{3}{c|}{CC3M} &
        \multicolumn{3}{c|}{COCO} &
        \multicolumn{3}{c}{OpenImages}                                                          \\
        \ method
                                  & R                         & P              & F$_1$
                                  & R                         & P              & F$_1$
                                  & R                         & P              & F$_1$          \\
        \shline
        \ CLIP
                                  & 0.799                     & 0.746          & 0.771
                                  & 0.774                     & 0.783          & 0.778
                                  & 0.762                     & 0.725          & 0.742          \\
        \ Flamingo
                                  & 0.842                     & \textbf{0.842} & 0.841
                                  & 0.835                     & \ul{0.922}     & 0.875
                                  & 0.838                     & \ul{0.863}     & 0.849          \\
        \ BLIP-2
                                  & 0.864                     & \ul{0.838}     & 0.850
                                  & 0.854                     & \textbf{0.961} & \ul{0.904}
                                  & 0.822                     & \textbf{0.864} & 0.841          \\
        \ InstBLIP
                                  & \ul{0.883}                & 0.827          & \ul{0.853}
                                  & \ul{0.892}                & 0.887          & 0.889
                                  & \ul{0.878}                & 0.842          & \textbf{0.859} \\
        \hline
        \ Ours
                                  & \textbf{0.908}            & 0.825          & \textbf{0.864}
                                  & \textbf{0.915}            & 0.911          & \textbf{0.913}
                                  & \textbf{0.881}            & 0.838          & \ul{0.858}
    \end{tabular}
    \vspace{-.7em}
    \caption{
        \textbf{Comparison with top-$10$ predictions} using CLIP ViT-L/14 \cite{radford2021learning} as the embedding model in evaluation metric.
    }
    \label{tab:results_ablation_on_embedding_models_in_evaluation}
    \vspace{-1.em}
\end{table}
%
%
\begin{table}[h]
    \tablestyle{1.8pt}{1.15}
    \begin{tabular}{x{32}|x{18.5}x{18.5}x{18.5}|x{18.5}x{18.5}x{18.5}|x{18.5}x{18.5}x{18.5}}
              & \multicolumn{3}{c|}{CC3M}
              & \multicolumn{3}{c|}{COCO}
              & \multicolumn{3}{c}{OpenImages}                 \\
        epoch & R                              & P     & F$_1$
              & R                              & P     & F$_1$
              & R                              & P     & F$_1$ \\
        \shline
        1     & 0.654                          & 0.487 & 0.553
              & 0.620                          & 0.623 & 0.620
              & 0.591                          & 0.520 & 0.548 \\
        2     & 0.698                          & 0.509 & 0.583
              & 0.659                          & 0.667 & 0.661
              & 0.604                          & 0.528 & 0.558 \\
        3     & 0.738                          & 0.530 & 0.611
              & 0.700                          & 0.712 & 0.702
              & 0.613                          & 0.544 & 0.570 \\
    \end{tabular}
    \vspace{-.7em}
    \caption{
        \textbf{Comparison of different training epochs} with top-$10$ predictions.
    }
    \label{tab:x_ablation_on_training_epochs}
    \vspace{-1.em}
\end{table}
%
%
\begin{table*}[t]
    \centering
    \begin{minipage}{1.\linewidth}
        \tablestyle{2.4pt}{1.15}
        \begin{tabular}{y{57}|y{94}|x{28}|z{36}|x{46}|x{18.5}x{18.5}x{18.5}|x{18.5}x{18.5}x{18.5}|x{18.5}x{18.5}x{17.5}}
                                      &                                                 &                &                        &               &
            \multicolumn{3}{c|}{CC3M} &
            \multicolumn{3}{c|}{COCO} &
            \multicolumn{3}{c}{OpenImages}                                                                                                                                \\
            method
                                      & \ models (vision + lang)                        & prompt         & data scale  \ \        & \# params (B)
                                      & R                                               & P              & F$_1$
                                      & R                                               & P              & F$_1$
                                      & R                                               & P              & F$_1$                                                          \\
            \shline
            \gray{CLIP} \cite{radford2021learning}
                                      & \ \gray{ViT L-14 + CLIP$_{\text{lang}}$}        & \gray{-}       & \gray{400M} \ \ \ \ \  & \gray{0.43}
                                      & \gray{0.515}                                    & \gray{0.481}   & \gray{0.493}
                                      & \gray{0.468}                                    & \gray{0.590}   & \gray{0.523}
                                      & \gray{0.460}                                    & \gray{0.485}   & \gray{0.467}                                                   \\
            \gray{CaSED} \cite{conti2023vocabulary}
                                      & \ \gray{ViT L-14 + Retrieval}                   & \gray{-}       & \gray{12M} \ \ \ \ \   & \gray{0.43}
                                      & \gray{0.577}                                    & \gray{0.520}   & \gray{0.541}
                                      & \gray{0.533}                                    & \gray{0.666}   & \gray{0.590}
                                      & \gray{0.490}                                    & \gray{0.506}   & \gray{0.492}                                                   \\
            \hline
            CLIP \cite{radford2021learning}
                                      & \ ViT L-14 + CLIP$_{\text{lang}}$               & -              & 400M \ \ \ \ \         & 0.43
                                      & 0.400                                           & 0.388          & 0.390
                                      & 0.385                                           & 0.489          & 0.427
                                      & 0.349                                           & 0.366          & 0.354                                                          \\
            CaSED \cite{conti2023vocabulary}
                                      & \ ViT L-14 + Retrieval                          & -              & 403M \ \ \ \ \         & 0.43          & 0.571 & 0.521 & 0.539
                                      & 0.532                                           & 0.683          & 0.596
                                      & 0.498                                           & 0.526          & 0.505                                                          \\
            Flamingo$_{\text{open}}$ \cite{awadalla2023openflamingo}
                                      & \ ViT L-14 + LLaMA 1 \cite{touvron2023llama}    & list           & 2.1B \ \ \ \ \         & 8.34
                                      & 0.542                                           & 0.541          & 0.535
                                      & 0.541                                           & 0.726          & 0.616
                                      & 0.524                                           & 0.622          & 0.561                                                          \\
            Flamingo$_{\text{open}}$
                                      & \ ViT L-14 + LLaMA 1                            & caption        & 2.1B \ \ \ \ \         & 8.34
                                      & 0.539                                           & 0.523          & 0.525
                                      & 0.547                                           & 0.712          & 0.614
                                      & 0.533                                           & 0.608          & 0.561                                                          \\
            Flamingo$_{\text{open}}$
                                      & \ ViT L-14 + MPT \cite{MosaicML2023Introducing} & list           & 2.1B \ \ \ \ \         & 8.13
                                      & 0.551                                           & 0.574          & 0.555
                                      & 0.552                                           & 0.801          & 0.648
                                      & 0.527                                           & \ul{0.657}     & 0.577                                                          \\
            Flamingo$_{\text{open}}$
                                      & \ ViT L-14 + MPT                                & caption        & 2.1B \ \ \ \ \         & 8.13
                                      & 0.532                                           & 0.537          & 0.528
                                      & 0.551                                           & 0.762          & 0.635
                                      & 0.544                                           & 0.655          & \ul{0.588}                                                     \\
            LLaVA$_\text{1.0}$ \cite{liu2023visual}
                                      & \ ViT L-14 + LLaMA 2 \cite{touvron2023llama2}   & list           & 753K \ \ \ \ \         & 13.3
                                      & 0.537                                           & 0.522          & 0.522
                                      & 0.574                                           & 0.790          & 0.659
                                      & 0.545                                           & 0.632          & 0.578                                                          \\
            LLaVA$_\text{1.0}$
                                      & \ ViT L-14 + LLaMA 2                            & caption        & 753K \ \ \ \ \         & 13.3
                                      & 0.588                                           & 0.520          & 0.547
                                      & 0.601                                           & 0.755          & 0.667
                                      & 0.545                                           & 0.557          & 0.545                                                          \\
            LLaVA$_\text{1.0}$
                                      & \ ViT L-14 + LLaMA 2                            & instruct       & 753K \ \ \ \ \         & 13.3
                                      & 0.566                                           & 0.507          & 0.531
                                      & 0.600                                           & 0.746          & 0.662
                                      & 0.567                                           & 0.589          & 0.571                                                          \\
            LLaVA$_\text{1.5}$ \cite{liu2023improvedllava}
                                      & \ ViT L-14 + Vicuna \cite{vicuna2023}           & list           & 1.2M \ \ \ \ \         & 13.4
                                      & 0.535                                           & 0.523          & 0.521
                                      & 0.581                                           & 0.800          & 0.666
                                      & 0.545                                           & 0.618          & 0.573                                                          \\
            LLaVA$_\text{1.5}$
                                      & \ ViT L-14 + Vicuna                             & caption        & 1.2M \ \ \ \ \         & 13.4
                                      & 0.581                                           & 0.510          & 0.543
                                      & 0.600                                           & 0.751          & 0.664
                                      & 0.551                                           & 0.560          & 0.555                                                          \\
            LLaVA$_\text{1.5}$
                                      & \ ViT L-14 + Vicuna                             & instruct       & 1.2M \ \ \ \ \         & 13.4
                                      & 0.552                                           & 0.530          & 0.532
                                      & 0.589                                           & 0.786          & 0.667
                                      & 0.566                                           & 0.607          & 0.576                                                          \\
            BLIP-2 \cite{li2023blip2}
                                      & \ ViT g-14 + Flant5xxl \cite{chung2022scaling}  & list           & 129M \ \ \ \ \         & 12.2
                                      & 0.541                                           & 0.558          & 0.541
                                      & 0.482                                           & 0.842          & 0.606
                                      & 0.466                                           & 0.626          & 0.526                                                          \\
            BLIP-2
                                      & \ ViT g-14 + Flant5xxl                          & caption        & 129M \ \ \ \ \         & 12.2
                                      & 0.594                                           & 0.549          & 0.564
                                      & 0.600                                           & \ul{0.894}     & 0.714
                                      & 0.523                                           & 0.626          & 0.561                                                          \\
            InstructBLIP \cite{dai2023instructblip}
                                      & \ ViT g-14 + Flant5xxl                          & list           & 129M \ \ \ \ \         & 12.3
                                      & 0.593                                           & 0.559          & 0.569
                                      & 0.613                                           & \textbf{0.897} & \ul{0.725}
                                      & 0.546                                           & 0.640          & 0.582                                                          \\
            InstructBLIP
                                      & \ ViT g-14 + Flant5xxl                          & caption        & 129M \ \ \ \ \         & 12.3
                                      & 0.603                                           & 0.535          & 0.561
                                      & 0.604                                           & 0.752          & 0.667
                                      & \ul{0.572}                                      & 0.585          & 0.572                                                          \\
            InstructBLIP
                                      & \ ViT g-14 + Flant5xxl                          & instruct       & 129M \ \ \ \ \         & 12.3
                                      & 0.529                                           & \textbf{0.605} & 0.556
                                      & 0.569                                           & 0.881          & 0.686
                                      & 0.559                                           & \textbf{0.698} & 0.614                                                          \\
            \hline
            Ours
                                      & \ ViT L-14 + Lang$_{\text{truncated}}$          & -              & 3M \ \ \ \ \           & 1.78
                                      & \textbf{0.673}                                  & \ul{0.598}     & \textbf{0.627}
                                      & \ul{0.618}                                      & 0.799          & 0.695
                                      & 0.560                                           & 0.595          & 0.570                                                          \\
            Ours
                                      & \ ViT L-14 + Lang$_{\text{truncated}}$          & -              & 70M \ \ \ \ \          & 1.78
                                      & \ul{0.659}                                      & 0.577          & \ul{0.609}
                                      & \textbf{0.674}                                  & 0.866          & \textbf{0.755}
                                      & \textbf{0.594}                                  & 0.615          & \textbf{0.597}                                                 \\
        \end{tabular}
    \end{minipage}
    \vspace{-.7em}
    \caption{
        \textbf{Comparison of different methods with top-$5$ predictions}.
        The same settings as in Table \ref{tab:main_results}.
    }
    \label{tab:x_main_results_top5}
    \vspace{-1.em}
\end{table*}
%
\subsection{Evaluation Metric}
\label{sec:x_evaluation_metric}
The recall in evaluation metric Eq. \ref{eq:pr} essentially represents the top-$k$ accuracy, which is for recognition tasks \cite{ILSVRC15}.
\noindentnewline
For an image, ground-truth (GT) labels are $\mathcal{G} = \{g_i\}_{i=1}^M$, ordered model predictions are $\mathcal{P} = \{p_j\}_{j=1}^N$.
The standard recall is defined as $R_{ecall} = TP / (TP + FN)$.
\noindentnewline
For recognition tasks, GT should either be TP (correctly identified) or FN (missed),
i.e., $TP + FN = |\mathcal{G}| = M$, then
\begin{align}
    R_{ecall} = \frac{TP}{TP + FN} = \frac{TP}{|\mathcal{G}|} = \frac{TP}{M}.
    \label{eq:recall}
\end{align}
\noindent
For closed-set recognition, $TP = \sum_{i=1}^M \mathbb{I}(g_i \in \mathcal{P})$,
where $g_i \in \mathcal{P}$ is a greedy matching -- correct prediction is exactly the same as $g_i$ with maximum semantic similarity,
e.g., $g_i = p_j =$ {\mytexttt{cat}}, and $\mathbb{I}(\cdot)$ is binary.
This $R_{ecall}$ is also called Exact Recall \cite{zhang2019bertscore}, also known as accuracy in image classification tasks \cite{ILSVRC15}.
In detail, to evaluate a classifier on ImageNet, each image has $M=1$ GT label and $N=1000$ class predictions, then Eq. \ref{eq:recall} becomes
\begin{align}
    \text{top-$k$ accuracy} = R_{ecall} = \mathbb{I}(g_1 \in \mathcal{P}_{1:k}),
\end{align}
For open-set recognition, $TP = \sum_{i=1}^M \mathbb{I}(g_i \in \mathcal{P})$, $g_i \in \mathcal{P}$ is a greedy matching but $\mathbb{I}(\cdot)$ is not binary because
correct prediction might not be exactly the same as $g_i$.
For instance, $g_i=$ {\mytexttt{cat}, $p_j=$ \mytexttt{kitty} or \mytexttt{feline} or \mytexttt{moggie}} are all correct with high semantic similarity,
and $p_j=$ {\mytexttt{dog} or \mytexttt{desk}} are wrong with low semantic similarity.
$\mathbb{I}(\cdot)$ is continuous to represent degrees of semantic similarity between $g_i$ and $p_j$.
One common choice for $\mathbb{I}(\cdot)$ is cosine similarity $\bS_{ij}$ between contextual embeddings of $g_i$ and $p_j$,
then Eq. \ref{eq:recall} becomes
\begin{align}
    R_{ecall} = \frac{1}{M} \sum_{i=1}^M {\max}_j \ \bS_{ij},
\end{align}
which is a.k.a. BERT Recall \cite{zhang2019bertscore}.
For the open-set case, each image has $M \geq 1$ GT labels and $N \geq 1$ predictions, then top-$k$ accuracy is
\begin{align}
    R_{ecall}^{top\text{-}k}
    = \frac{1}{M} \sum_{i=1}^M \mathbb{I}(g_i \in \mathcal{P}_{1:k})
    = \frac{1}{M} \sum_{i=1}^M {\max}_{j \in [1, k]} \ \mathbf{S}_{ij}.
\end{align}
The top-$k$ refers to the $k$ most relevant predictions of \textit{all possible labels in the world} to the image.
\subsection{Data Preprocessing}
\label{sec:x_data_preprocessing}
\noindent
For an image, the paired caption is preprocessed using the steps summarized in the following table.
%
\begin{center}
    \tablestyle{2.pt}{1.15}
    \begin{tabular}{c|y{215}}
        step & \ details                                                                          \\
        \shline
        1    & \ Lowercase the caption.                                                           \\
        2    & \ Eliminate high-frequency noise words that lack meaningful                        \\
             & \ content. The noise words removed in our work are $[$ person,                     \\
             & \ persons, stock, image, images, background, ounce, illustration,                  \\
             & \ front, photography, day $]$.                                                     \\
        3    & \ Keep only the letters, and a few special characters like spaces ( ),             \\
             & \ periods (.), commas (,), ampersands (\&), and hyphens (-).                       \\
             & \ Exclude all others, including numbers and words containing                       \\
             & \ numbers.                                                                         \\
        4    & \ Use NLTK \cite{bird2009natural} to tokenize the caption into words. Then tag the \\
             & \ words with their part-of-speech (POS) tags to filter out words that              \\
             & \ are not nouns. The noun tags used in this paper are $[$ NN, NNS $]$.             \\
        5    & \ Lemmatize the words to their root forms. For example, the word                   \\
             & \ ``dogs'' is lemmatized to ``dog''.                                               \\
    \end{tabular}
\end{center}
%
With this preprocessing, we obtain a set of meaningful noun words for each image and summarize the information in the following table, including the number of image-caption pairs and distinct nouns.
%
\begin{center}
    \tablestyle{2.pt}{1.15}
    \begin{tabular}{l|x{21}x{20}|x{24}x{17}|x{26}|x{42}|x{24}}
                                        &
        \multicolumn{2}{c|}{CC3M}       &
        \multicolumn{2}{c|}{COCO}       &
        \multicolumn{1}{c|}{SBU}        &
        \multicolumn{1}{c|}{OpenImages} &
        \multicolumn{1}{c}{{L{\scriptsize{AION}}}}                                                    \\
        statistics                      &
        train                           & val   & train    & val     & train &
        val                             & train                                                       \\
        \shline
        \# images                       & 2.69M & 12478    & 118287  & 5000  & 828816 & 41686 & \ 67M \\
        \# nouns                        & 22890 & \ \ 4875 & \ 15444 & 3834  & 132372 & 3119  & 2.7M  \\
    \end{tabular}
\end{center}
%
The training split contains 2,794,419 distinct nouns, while all validation splits have a total of 8,637 distinct nouns.
The number of overlapping nouns between the training and validation splits is 8,347, which is 97.8\% of distinct nouns in validation splits.
\subsection{Prompt Settings}
\label{sec:x_prompt_settings}
\noindent
For training, we adopt the prompt augmentation, which contains different prompt templates but with the same semantic meaning.
In each training iteration, we randomly select one prompt from those templates for the batched images.
For inference, we only use one simple prompt in all experiments.
The prompt templates are listed as follows.
%
\begin{center}
    \tablestyle{2.pt}{1.15}
    \begin{tabular}{y{32}|y{140}}
        setting   & \ prompt templates                           \\
        \shline
        training  & \ The objects in the image are               \\
                  & \ The items present in the picture are       \\
                  & \ The elements depicted in the image are     \\
                  & \ The objects shown in the photograph are    \\
                  & \ The items visible in the image are         \\
                  & \ The objects that appear in the picture are \\
                  & \ The elements featured in the image are     \\
                  & \ The items captured in the photograph are   \\
                  & \ The elements seen in the picture are       \\
                  & \ The items represented in the image are     \\
        \hline
        inference & \ The objects in the image are
    \end{tabular}
\end{center}
%
For comparison, we evaluate chat-based VQA models, i.e., BLIP-2 \cite{li2023blip2}, InstructBLIP \cite{dai2023instructblip}, and LLaVA \cite{liu2023visual,liu2023improvedllava}, with two types of prompt, which are
\begin{enumerate}
    \item[1)] text completion: {\mytexttt{The objects in the image are}},
    \item[2)] and VQA: {\mytexttt{Describe every detail in the image.}}
\end{enumerate}
We refer to the text completion prompt as \textit{prompt: list} and the VQA prompt as \textit{prompt: caption}.
After obtaining model outputs, we apply the rule from Section \ref{sec:x_data_preprocessing} to extract nouns as predicted labels.
\noindentnewline
Especially, Flamingo \cite{alayrac2022flamingo,awadalla2023openflamingo} has a unique prompt setting with few-shot instruction.
For the \textit{caption} type, we change the prompt setting to {\mytexttt{What objects are in the image?}}.
Then we construct the prompt with 4-shot samples as in \cite{alayrac2022flamingo}, which is listed as the following tables.
%
\begin{center}
    \tablestyle{2.pt}{1.15}
    \begin{tabular}{y{233}}
        the \textit{list} prompt type with few-shot samples for Flamingo \\
        \shline
        \mytexttt{<}image\mytexttt{>}The objects in the image are boy, bush, chair, clothes, grass, house, tree, sports ball.\mytexttt{<}$|$endofchunk$|$\mytexttt{>}
        \mytexttt{<}image\mytexttt{>}The objects in the image are bus, car, clouds, house, leaves, person, road.\mytexttt{<}$|$endofchunk$|$\mytexttt{>}
        \mytexttt{<}image\mytexttt{>}The objects in the image are giraffe, grass, tree.\mytexttt{<}$|$endofchunk$|$\mytexttt{>}
        \mytexttt{<}image\mytexttt{>}The objects in the image are cat, telecontroller, sofa.\mytexttt{<}$|$endofchunk$|$\mytexttt{>}
        \mytexttt{<}image\mytexttt{>}The objects in the image are        \\
        \hline
        \vspace{.1em}
        the reference images as few-shot samples for Flamingo            \\
        \shline
    \end{tabular}
    \centering
    \includegraphics[width=1\linewidth]{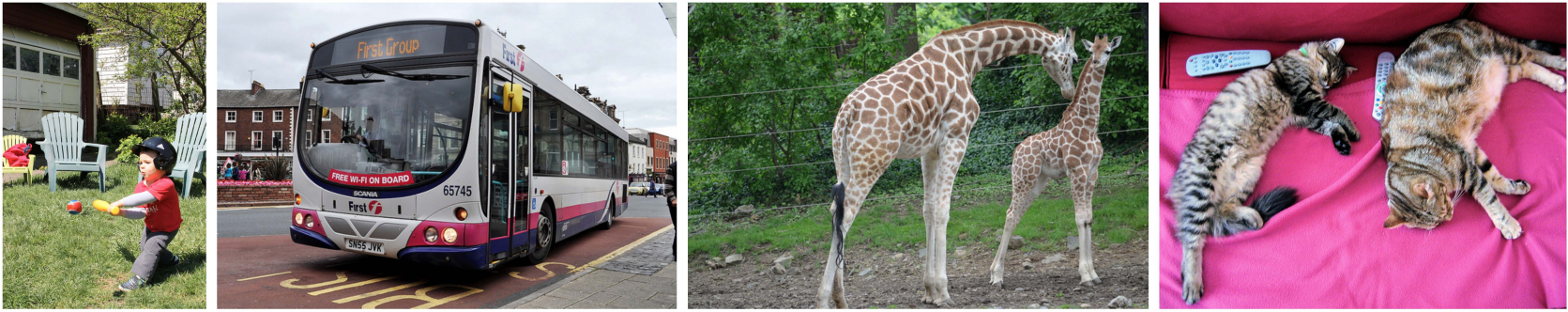}
\end{center}
%
\subsection{Number of Sampling Tokens in Comparison}
\noindent
We have various models to compare with ours.
For a fair comparison, we need to take care of the maximum number of sampling tokens for each model to make sure that we can extract enough potential nouns words from their outputs.
LLaVA \cite{liu2023visual,liu2023improvedllava} has a maximum number of sampling tokens of 1024, which is already enough for the task.
BLIP-2 \cite{li2023blip2} has a maximum 32 in default, but we change it to 64 for top-$5$ and 128 for top-$10$.
To verify this setting is fair for BLIP-2, we ablate the number of sampling tokens for BLIP-2 with the caption prompt in Table \ref{tab:x_ablation_on_number_of_sampling_tokens_for_blip2}.
%
\begin{table}[t]
    \tablestyle{1.8pt}{1.15}
    \begin{tabular}{x{32}|x{18.5}x{18.5}x{18.5}|x{18.5}x{18.5}x{18.5}|x{18.5}x{18.5}x{18.5}}
                                  &
        \multicolumn{3}{c|}{CC3M} &
        \multicolumn{3}{c|}{COCO} &
        \multicolumn{3}{c}{OpenImages}                    \\
        \# tokens                 & R     & P     & F$_1$
                                  & R     & P     & F$_1$
                                  & R     & P     & F$_1$ \\
        \shline
        \multicolumn{10}{l}{\textit{prompt: list}}        \\
        \hline
        64                        & 0.542 & 0.556 & 0.540
                                  & 0.482 & 0.842 & 0.606
                                  & 0.455 & 0.622 & 0.518 \\
        128                       & 0.544 & 0.557 & 0.542
                                  & 0.494 & 0.871 & 0.623
                                  & 0.476 & 0.641 & 0.538 \\
        256                       & 0.542 & 0.556 & 0.540
                                  & 0.482 & 0.842 & 0.606
                                  & 0.455 & 0.622 & 0.518 \\
        \hline
        \multicolumn{10}{l}{\textit{prompt: caption}}     \\
        \hline
        64                        & 0.601 & 0.539 & 0.561
                                  & 0.600 & 0.893 & 0.714
                                  & 0.523 & 0.626 & 0.562 \\
        128                       & 0.609 & 0.539 & 0.561
                                  & 0.600 & 0.893 & 0.714
                                  & 0.523 & 0.626 & 0.562 \\
        256                       & 0.600 & 0.539 & 0.560
                                  & 0.601 & 0.894 & 0.714
                                  & 0.512 & 0.643 & 0.562 \\
    \end{tabular}
    \vspace{-.7em}
    \caption{\textbf{Different number of sampling tokens} for BLIP-2 with top-$10$ predictions.}
    \label{tab:x_ablation_on_number_of_sampling_tokens_for_blip2}
    \vspace{-1.em}
\end{table}
%
For InstructBLIP \cite{dai2023instructblip}, we use its default number of sampling tokens, which is 256 for top-$5$ and top-$10$.
To verify the setting, we ablate the number of sampling tokens for InstructBLIP in Table \ref{tab:x_ablation_on_number_of_sampling_tokens_for_instructblip}.
%
\begin{table}[t]
    \tablestyle{1.8pt}{1.15}
    \begin{tabular}{x{32}|x{18.5}x{18.5}x{18.5}|x{18.5}x{18.5}x{18.5}|x{18.5}x{18.5}x{18.5}}
                                  &
        \multicolumn{3}{c|}{CC3M} &
        \multicolumn{3}{c|}{COCO} &
        \multicolumn{3}{c}{OpenImages}                    \\
        \# tokens                 & R     & P     & F$_1$
                                  & R     & P     & F$_1$
                                  & R     & P     & F$_1$ \\
        \shline
        \multicolumn{10}{l}{\textit{prompt: list}}        \\
        \hline
        256                       & 0.596 & 0.554 & 0.567
                                  & 0.613 & 0.897 & 0.725
                                  & 0.546 & 0.640 & 0.582 \\
        512                       & 0.596 & 0.554 & 0.567
                                  & 0.613 & 0.897 & 0.725
                                  & 0.544 & 0.634 & 0.578 \\
        \hline
        \multicolumn{10}{l}{\textit{prompt: caption}}     \\
        \hline
        256                       & 0.639 & 0.487 & 0.546
                                  & 0.690 & 0.662 & 0.673
                                  & 0.647 & 0.539 & 0.581 \\
        512                       & 0.639 & 0.487 & 0.546
                                  & 0.690 & 0.662 & 0.673
                                  & 0.647 & 0.539 & 0.581 \\
    \end{tabular}
    \vspace{-.7em}
    \caption{\textbf{Different number of sampling tokens} for InstructBLIP with top-$10$ predictions.}
    \label{tab:x_ablation_on_number_of_sampling_tokens_for_instructblip}
    \vspace{-1.em}
\end{table}
%
Due to Flamingo \cite{alayrac2022flamingo,awadalla2023openflamingo} has the same output format as ours, we keep the same maximum number of sampling tokens for it as ours for greedy search, i.e., 64 for top-$5$.
We double the number to 128 for its top-$10$ predictions.
For VQA methods, sampling more tokens for more potential predictions significantly increases time cost, esp. with beam search.
\subsection{Visualizing Predictions}
\label{sec:x_results_visualization}
\begin{figure*}[t]
    \centering
    \includegraphics[width=1.\linewidth]{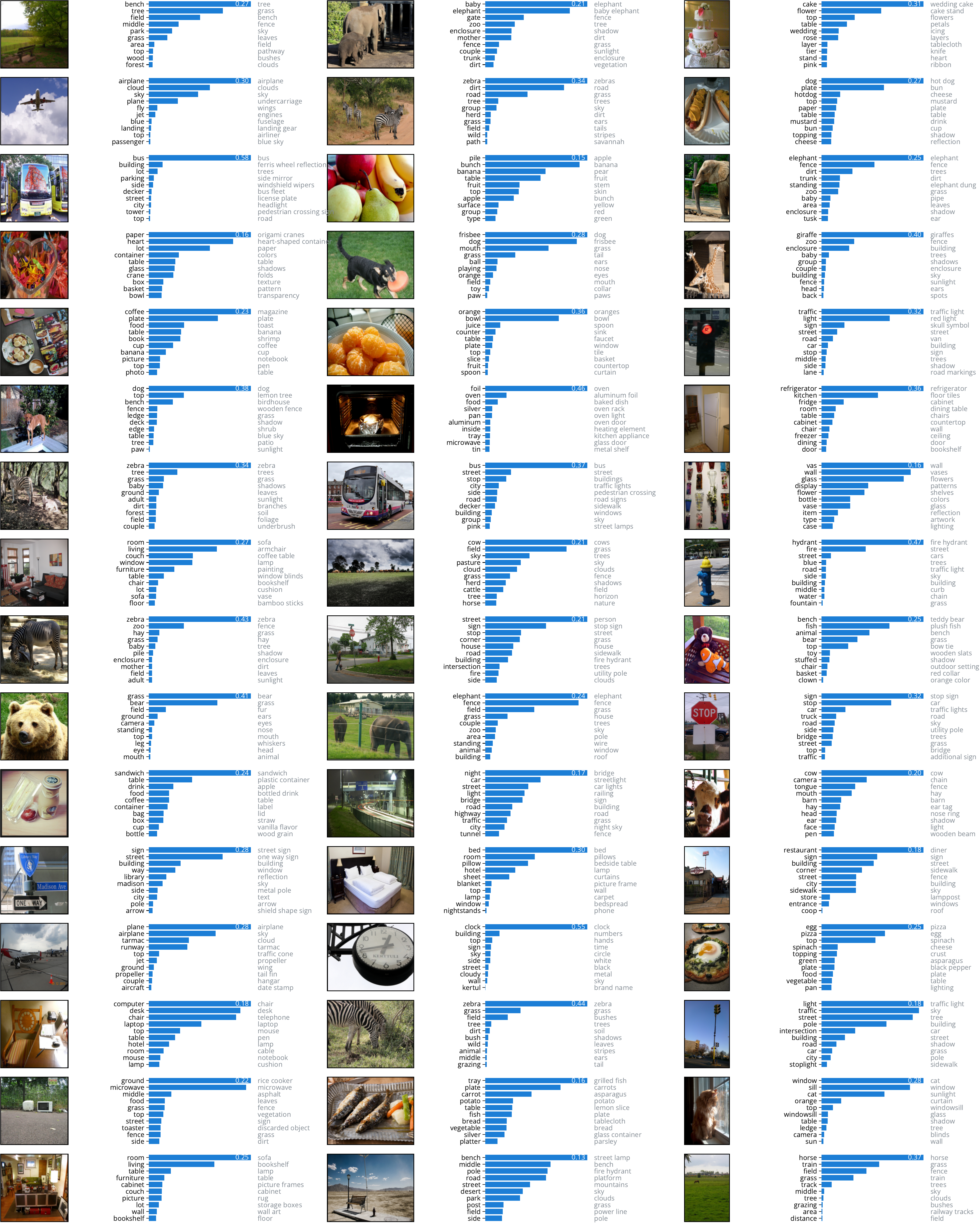}
    \caption{
        Top-$10$ predictions on COCO validation split without cherry-picking.
        The top bar is with the first prediction's probability.
        The right column shows predictions \gray{in gray} from the GPT-4V Preview.
        Images are licensed under a \href{https://www.flickr.com/creativecommons/}{Creative Commons Attribution 2.0 License}.
    }
    \label{fig:x_ours_results_A}
\end{figure*}

\begin{figure*}[t]
    \centering
    \includegraphics[width=1.\linewidth]{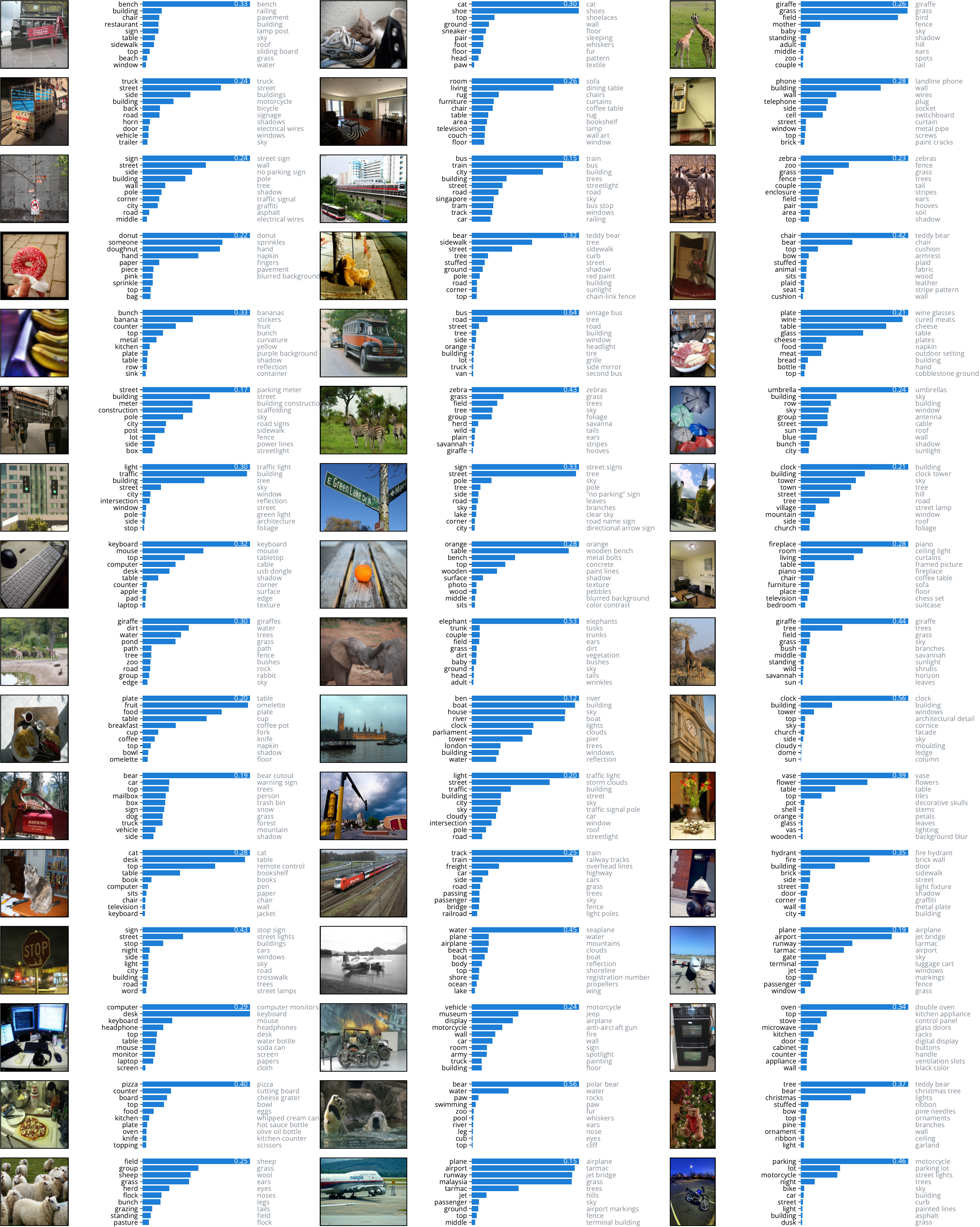}
    \caption{
        Top-$10$ predictions on COCO validation split without cherry-picking.
        The top bar is with the first prediction's probability.
        The right column shows predictions \gray{in gray} from the GPT-4V Preview.
        Images are licensed under a \href{https://www.flickr.com/creativecommons/}{Creative Commons Attribution 2.0 License}.
    }
    \label{fig:x_ours_results_B}
\end{figure*}

\begin{figure*}[t]
    \centering
    \includegraphics[width=1.\linewidth]{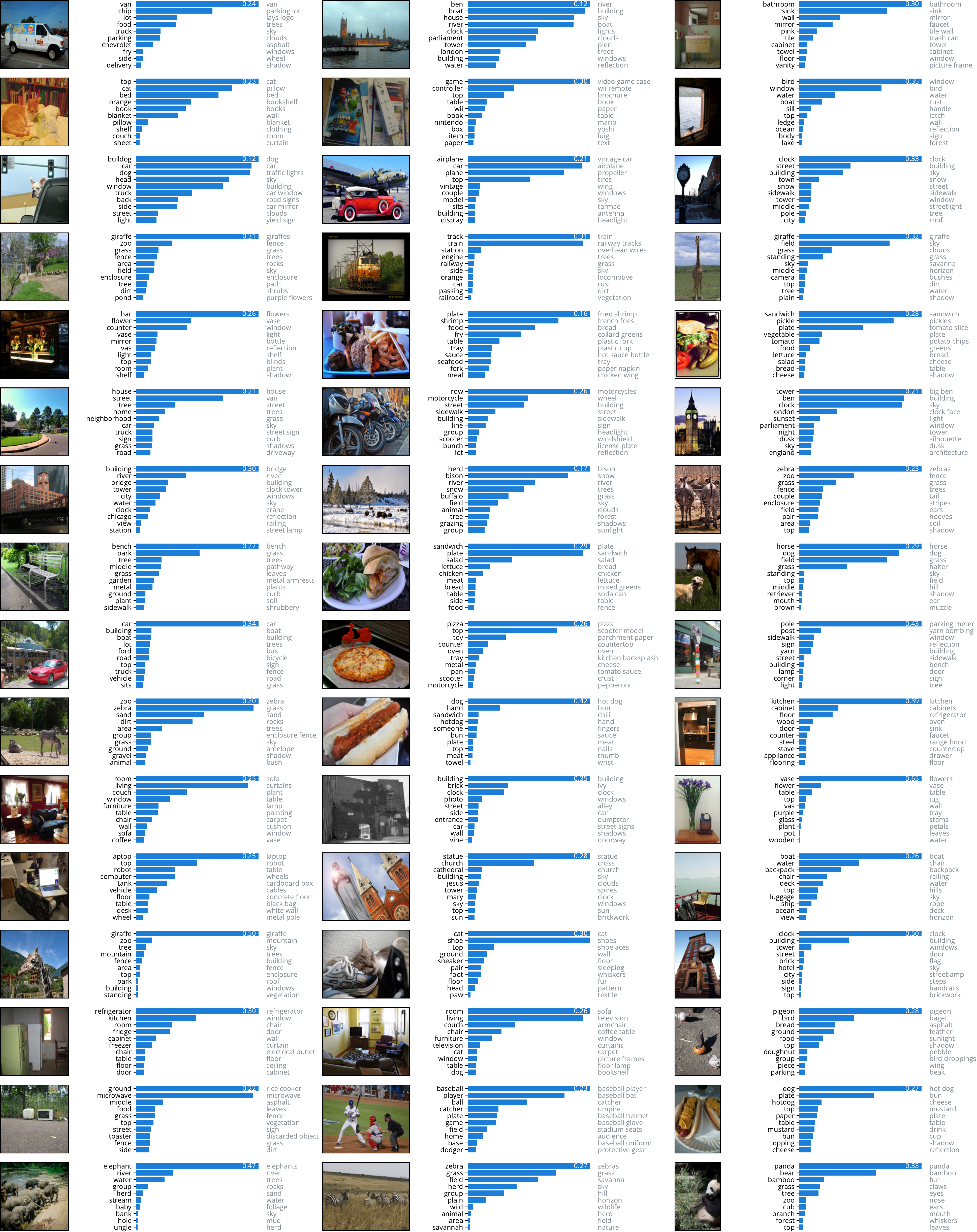}
    \caption{
        Top-$10$ predictions on COCO validation split without cherry-picking.
        The top bar is with the first prediction's probability.
        The right column shows predictions \gray{in gray} from the GPT-4V Preview.
        Images are licensed under a \href{https://www.flickr.com/creativecommons/}{Creative Commons Attribution 2.0 License}.
    }
    \label{fig:x_ours_results_C}
\end{figure*}

\begin{figure*}[t]
    \centering
    \includegraphics[width=1.\linewidth]{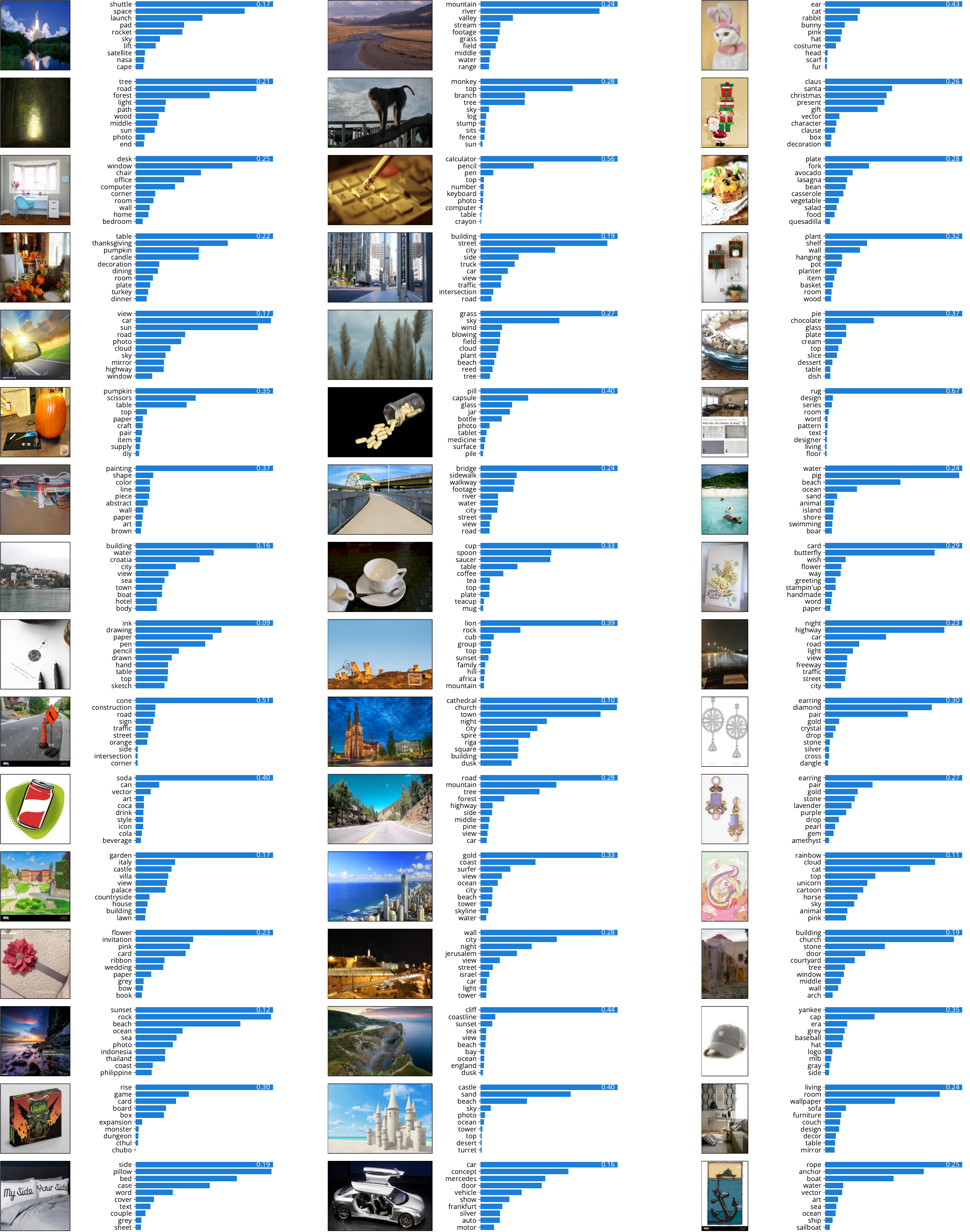}
    \caption{
        Top-$10$ predictions on CC3M validation split without cherry-picking.
        The top bar is with the first prediction's probability.
        Images in the dataset of CC3M are \href{https://github.com/google-research-datasets/conceptual-captions/blob/master/LICENSE}{provided by Google LLC}.
    }
    \label{fig:x_ours_results_D}
\end{figure*}

\begin{figure*}[t]
    \centering
    \includegraphics[width=1.\linewidth]{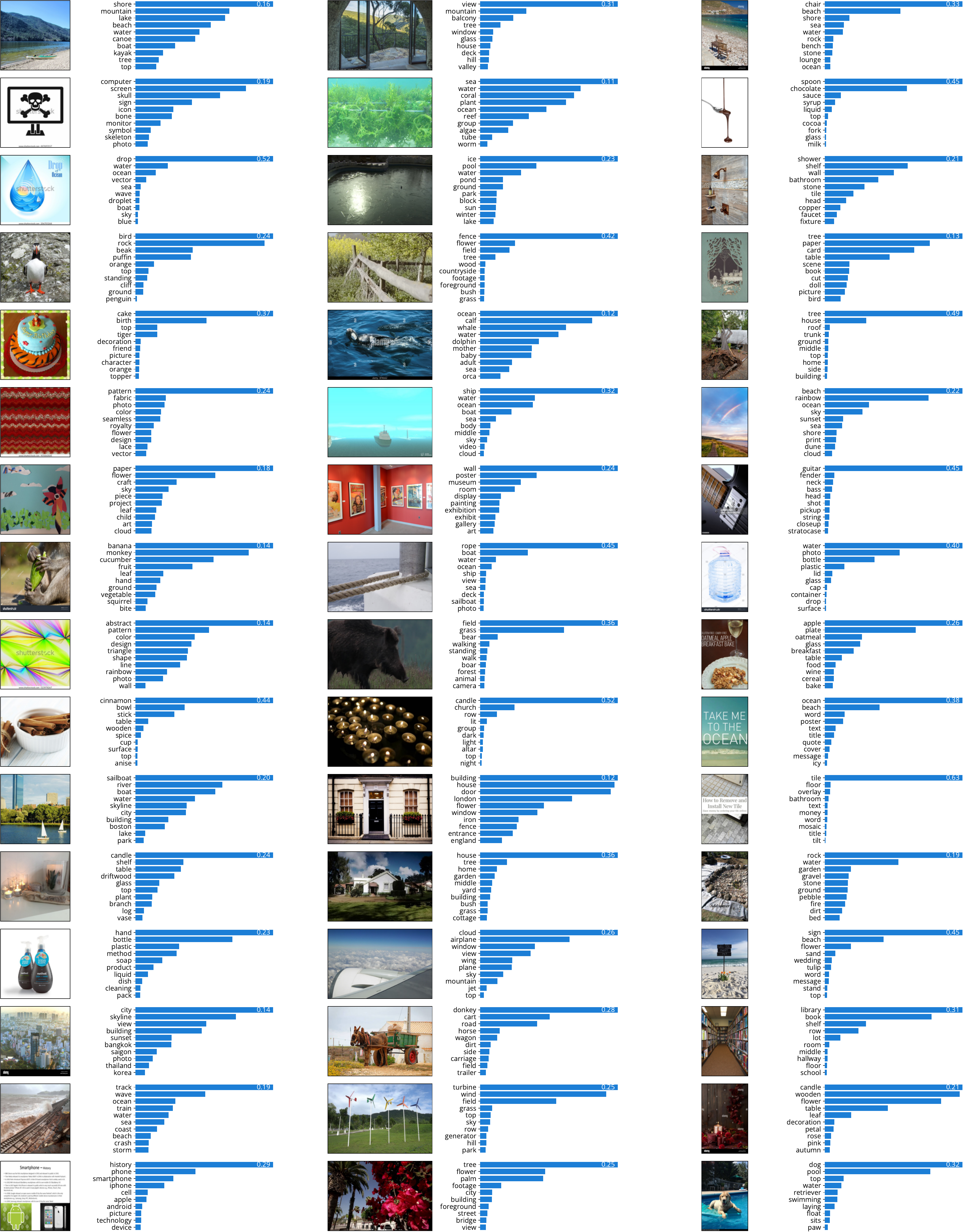}
    \caption{
        Top-$10$ predictions on CC3M validation split without cherry-picking.
        The top bar is with the first prediction's probability.
        Images in the dataset of CC3M are \href{https://github.com/google-research-datasets/conceptual-captions/blob/master/LICENSE}{provided by Google LLC}.
    }
    \label{fig:x_ours_results_E}
\end{figure*}

\begin{figure*}[t]
    \centering
    \includegraphics[width=1.\linewidth]{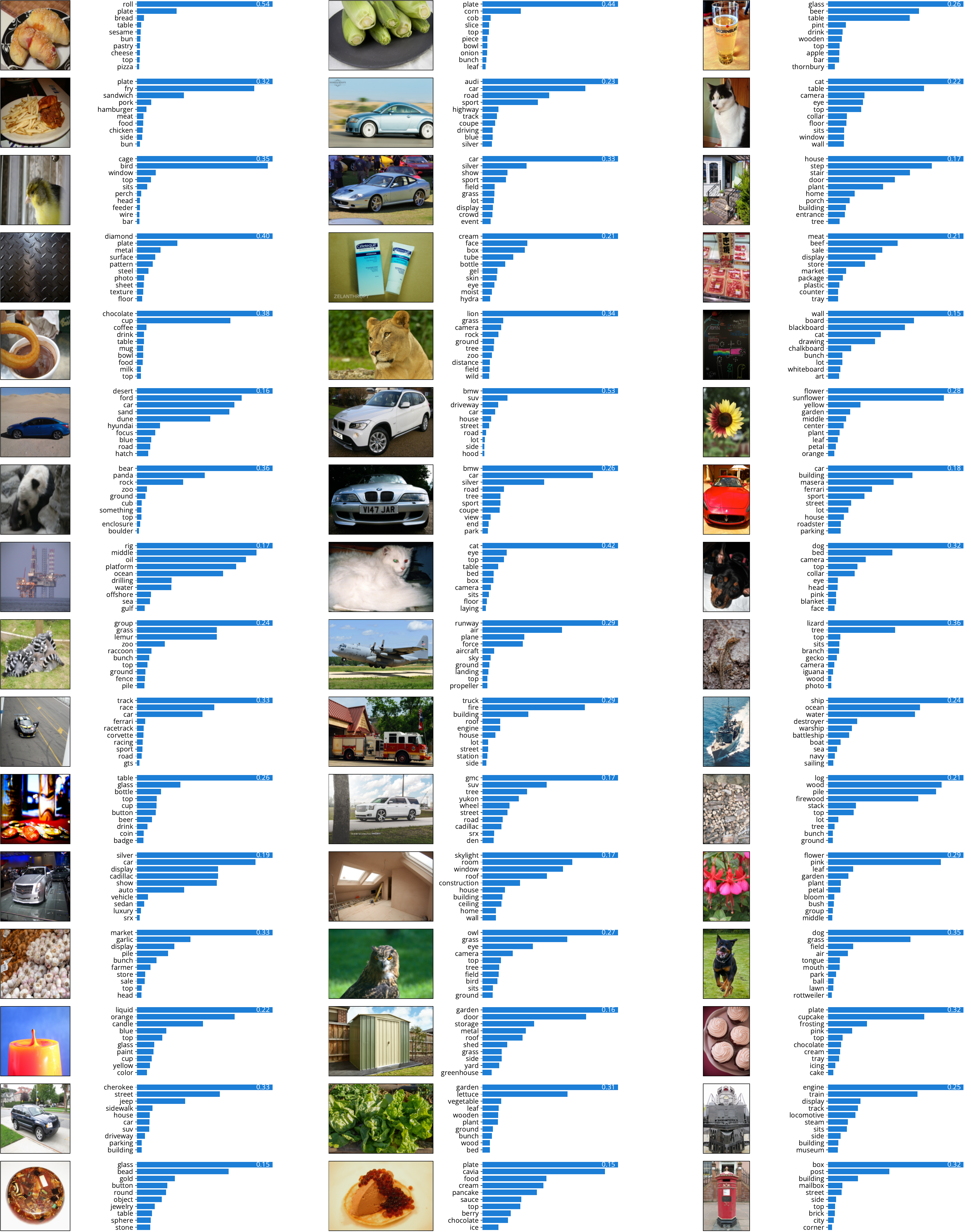}
    \caption{
        Top-$10$ predictions on OpenImages validation split without cherry-picking.
        The top bar is with the first prediction's probability.
        Images in the dataset of OpenImages are under a
        \href{https://storage.googleapis.com/openimages/web/factsfigures_v7.html}{Creative Commons Attribution 2.0 License}.
    }
    \label{fig:x_ours_results_F}
\end{figure*}

\begin{figure*}[t]
    \centering
    \includegraphics[width=1.\linewidth]{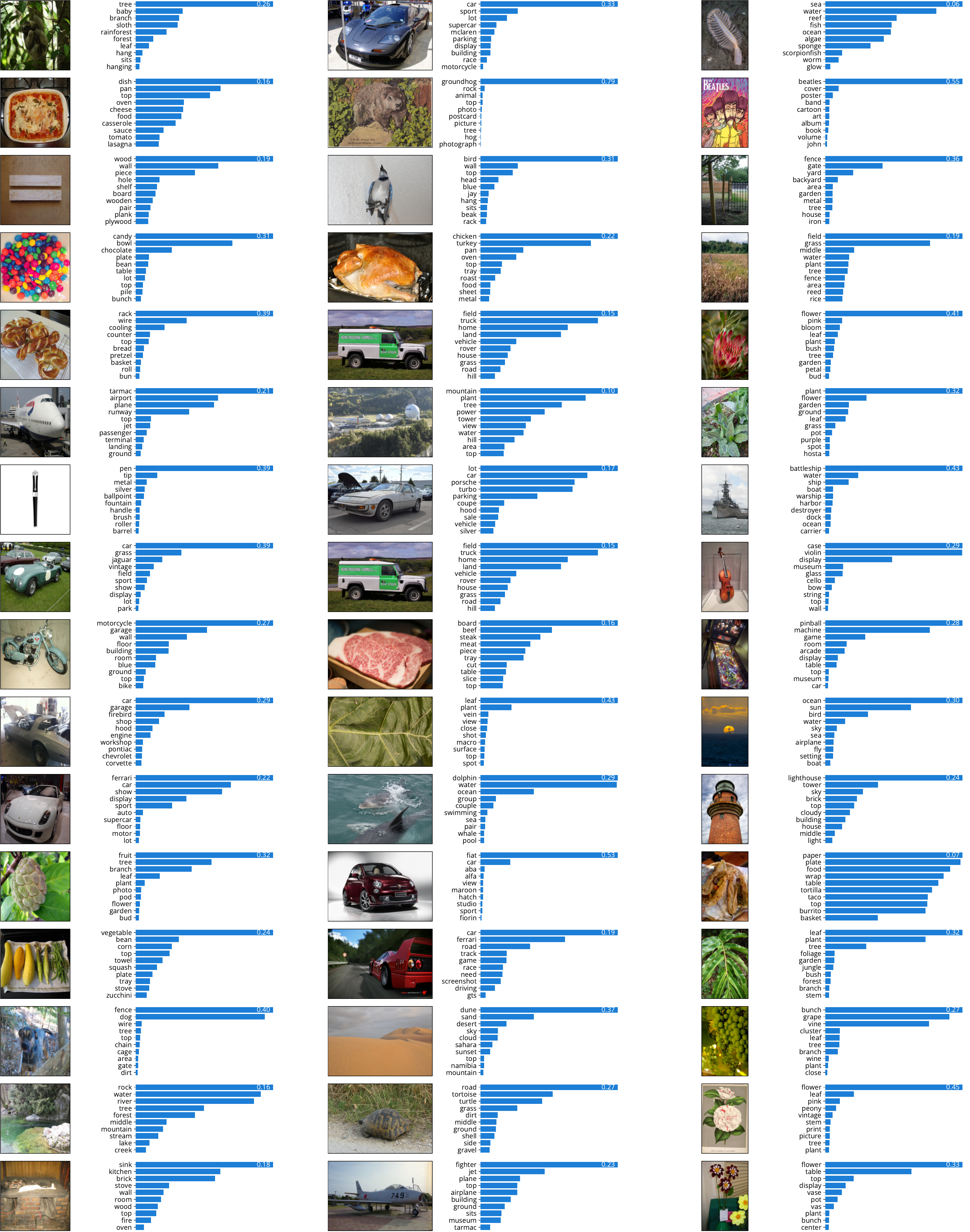}
    \caption{
        Top-$10$ predictions on OpenImages validation split without cherry-picking.
        The top bar is with the first prediction's probability.
        Images in the dataset of OpenImages are under a
        \href{https://storage.googleapis.com/openimages/web/factsfigures_v7.html}{Creative Commons Attribution 2.0 License}.
    }
    \label{fig:x_ours_results_G}
\end{figure*}

\noindent
We visualize the top-$10$ predictions from our 1.78B model trained on G70M in Figure \ref{fig:x_ours_results_A}-\ref{fig:x_ours_results_G} without cherry-picking.
The image is paired with two columns: our predictions on the left, probability-indicating ranking bars on the right.
The images sampled from COCO have \textcolor{mygray}{gray} column to show GPT-4V Preview's \cite{gpt-4v} predictions, intuitively illustraing the strengths and weaknesses of our method with the apples-to-apples comparison.
\subsection{Discussion}
\label{sec:x_discussion}
\noindent
In this section, we discuss the limitations of our method and experiments that we have tried but does not work well.
\noindentnewline
\textbf{Less Is More}.
Our method's performance heavily relies on the quality of the training data.
More noisy data will hurt the performance, for example, models trained on the noisier CC12M \cite{changpinyo2021conceptual} underperform compared to those trained on CC3M \cite{sharma2018conceptual}.
Moreover, high quality requires more human efforts, which is expensive, meaning to densely annotate all possible labels for each image.
We might consider using GPT-4V \cite{gpt-4v} for generating high-quality labels, though it may be costly (API expenses) and subject to the hallucination issue.
Exploring methods to train models with fewer labels for broader generalization could be intriguing.
\noindentnewline
\textbf{Defining Labels}.
How to define the label for describing an object in an image?
A label could be a word, which is used in this paper, but also could be a phrase or a sentence.
We have tried to define the label with the noun phrase, which includes an adjective, for example, ``gray car'' and ``cute boy''.
However, these models underperformed, partly due to poor training data quality and the limitations of the parser for extracting noun phrases from captions.
We also experimented with concrete nouns for training, but the results were unsatisfactory due to noisy reference labels produced by the parser, which needs a comprehensive filter to remove noise.
\noindentnewline
\textbf{Evaluation}.
First, our evaluation has limitations due to the incomplete and imperfect nature of reference labels derived from raw captions.
Second, we calculate $P$, $R$ and $F_1$ score based on the semantic similarity between the embeddings of predicted and reference labels from a pretrained language model.
However, such a model-based semantic similarity brings noise and bias to the evaluation results due to the model imperfection.
This motivates us to conduct the cross-validation experiments in Section \ref{sec:x_compare_with_gpt4visionpreview}, which views GPT-4V's \cite{gpt-4v} predictions as reference labels.
Developing a reliable evaluation metric beyond human evaluation or model-based semantic similarity is an interesting topic.
\noindentnewline
\textbf{Fine-Grained Recognition}.
Our method, though not designed for fine-grained recognition, could be adapted for such tasks.
Currently, the method underperforms in this area due to the use of general, rather than fine-grained, training data.
Improving performance may be possible by using more specific, fine-grained training data, which circles back to the initial question regarding the quality of training data.
\noindentnewline
\textbf{Single-Label Prediction}.
Our method is optimized for top-$k$ predictions and exhibits lower performance in top-$1$ accuracy evaluations.
Our approach encourages the model to predict multiple labels for an image, which is more realistic than predicting just one label because images generally contain multiple objects.
Therefore, we do not focus on improving top-$1$ accuracy in this paper.
\noindentnewline
\textbf{Competition Issue}.
We acknowledge the inherent competitive issue in our one-shot sampling, similar to the repetition issue observed in sequence-based methods like greedy and beam search.
However, its results are still promising in experiments, which is likely due to redundant tokenization.
Mitigating or analyzing the competition issue for the one-shot sampling could be our future research topic.
\subsection{Other Related Works}
\label{sec:x_other_related_works}
\noindent
Approaching object recognition as a natural language prediction, pioneered by \cite{mori1999image,barnard2001learning,duygulu2002object}, has been proposed before the deep learning era \cite{lecun2015deep}.
The motivation is primarily to assist journalists in annotating images for retrieval purposes \cite{markkula2000end,barnard2003matching}.
\cite{mori1999image} slices an image into regions and predicts words using probabilistic models.
\cite{duygulu2002object} views recognition as a machine translation problem, aligning image regions with words using a lexicon, optimized by the EM algorithm \cite{dempster1977maximum}.
\noindentnewline
\textbf{Image Annotation and Multi-label Prediction}.
The evolution of image annotation or tagging closely mirrors that of multi-label prediction.
Initial approaches develop on topic models \cite{jia2011learning} like latent Dirichlet allocation \cite{barnard2003matching} and probabilistic latent semantic analysis \cite{hofmann2001plsa,monay2004plsa}.
Mixture models \cite{jeon2003automatic,lavrenko2003model,feng2004multiple} have also been explored to model the joint distributions over images and tags.
Then SVM-based discriminative models \cite{joachims2002optimizing,cusano2003image,hertz2004learning} are proposed to predict tags.
Later, the annotation task is treated as a retrieval problem \cite{makadia2008new,guillaumin2009tagprop} based on nearest neighbors \cite{cover1967nearest} or joint optimization \cite{chen2013fast}.
The difficulty of collecting multi-label annotations inspires curriculum learning-based models \cite{durand2019learning,cole2021multi} and semi-supervised methods \cite{fergus2009semi,schroff2010harvesting,socher2010connecting}.
Now models with ranking-based losses \cite{gong2013deep} and transformer-based architecture \cite{liu2021query2label,huang2023tag2text,zhang2023recognize,ridnik2023ml} are introduced for tagging images, but they are still closed-set recognition models trained on heavily-annotated/cleaned datasets.
In contrast, our method is an open-set recognition model trained on raw data, which is at the real open-level with a large-scale prediction capability (top-$100$).
In the figure below, our model correctly predicts the wild terms such as {\mytexttt{sora}, \mytexttt{cloudscape}, \mytexttt{text}, \mytexttt{logo}, \mytexttt{letter}, \mytexttt{art}, and \mytexttt{animation}}, assigning probabilities for ranking or filtering, while \cite{zhang2023recognize} does not.
%
\begin{center}
    \includegraphics[width=1\linewidth]{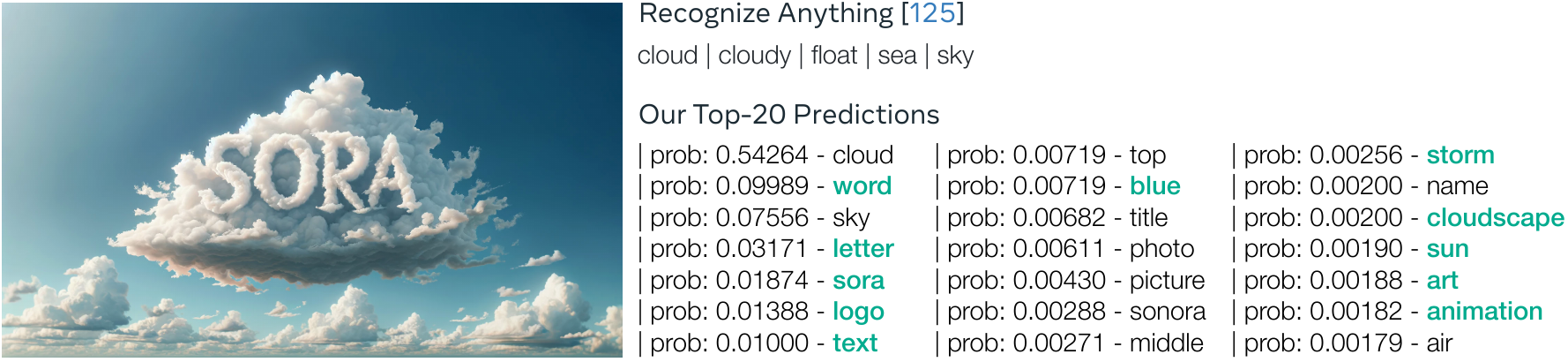}
\end{center}
%
\subsection{Acknowledgements}
\label{sec:acknowledgements}
\noindent
We thank \href{https://scholar.google.com/citations?user=EPImyCcAAAAJ&hl=en}{\textcolor{black}{Alessandro Conti}}, the primary author of CaSED \cite{conti2023vocabulary}, for supplying the text embedding galleries for CC3M, COCO, SBU, and LAION-400M.
We also thank Damian Gessler for the help on downloading training datasets and solving cluster issues, and our group colleagues at Meta for the helpful discussions.
%
\noindent
%